\documentclass[final]{cvpr}

\usepackage{lipsum}

\newcommand\blfootnote[1]{%
  \begingroup
  \renewcommand\thefootnote{}\footnote{#1}%
  \addtocounter{footnote}{-1}%
  \endgroup
}

\usepackage{times}
\usepackage{epsfig}
\usepackage{graphicx}
\usepackage{amsmath}
\usepackage{amssymb}
\usepackage{caption}

\usepackage[pagebackref=true,breaklinks=true,colorlinks,bookmarks=false]{hyperref}

\def\cvprPaperID{6565} % *** Enter the CVPR Paper ID here
\def\confYear{CVPR 2021}

\begin{document}
%\pagenumbering{gobble}
\linespread{0.94}
%%%%%%%%% TITLE
\title{Normalized Avatar Synthesis Using StyleGAN and Perceptual Refinement}

\author{Huiwen Luo\hspace{0.3in} Koki Nagano \hspace{0.3in}Han-Wei Kung \hspace{0.3in}Mclean Goldwhite
\hspace{0.3in} Qingguo Xu 
\vspace{0.01pt}
\and
Zejian Wang \hspace{0.3in}  Lingyu Wei \hspace{0.3in} Liwen Hu \hspace{0.3in} Hao Li
\vspace{0.1in}
\\
Pinscreen
\vspace{-0.15in}
}

\twocolumn[{
\renewcommand\twocolumn[1][]{#1}%
\maketitle
\begin{center}
    \centering
    \includegraphics[width=6.8in]{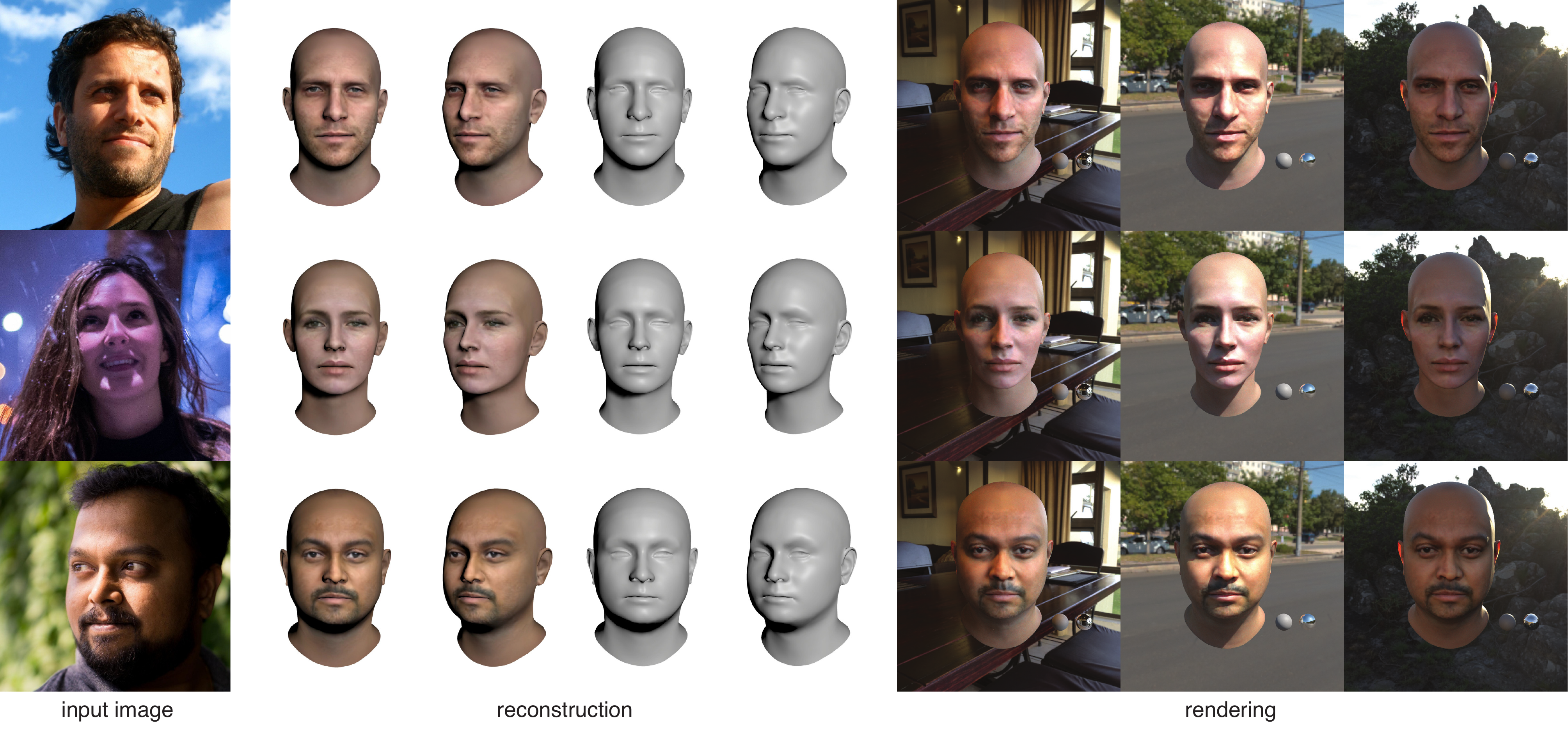}
    \captionof{figure}{Given a single photograph, we can reconstruct a high-quality textured 3D face with neutral expression and normalized lighting condition. Our approach can handle extremely challenging cases and our generated avatars are animation friendly and suitable for complex relighting in virtual environments.}
    \label{fig:teaser}
\end{center}
}]

\blfootnote{Hao Li is affiliated with Pinscreen and UC Berkeley; Koki Nagano is currently at NVIDIA. This work was fully conducted at Pinscreen.}
%%%%%%%%% ABSTRACT
\begin{abstract}

We introduce a highly robust GAN-based framework for digitizing a normalized 3D avatar of a person from a single unconstrained photo. While the input image can be of a smiling person or taken in extreme lighting conditions, our method can reliably produce a high-quality textured model of a person's face in neutral expression and skin textures under diffuse lighting condition. Cutting-edge 3D face reconstruction methods use non-linear morphable face models combined with GAN-based decoders to capture the likeness and details of a person but fail to produce neutral head models with unshaded albedo textures which is critical for creating relightable and animation-friendly avatars for integration in virtual environments. The key challenges for existing methods to work is the lack of training and ground truth data containing normalized 3D faces. We propose a two-stage approach to address this problem. First, we adopt a highly robust normalized 3D face generator by embedding a non-linear morphable face model into a StyleGAN2 network. This allows us to generate detailed but normalized facial assets. This inference is then followed by a perceptual refinement step that uses the generated assets as regularization to cope with the limited available training samples of normalized faces. We further introduce a Normalized Face Dataset, which consists of a combination photogrammetry scans, carefully selected photographs, and generated fake people with neutral expressions in diffuse lighting conditions. While our prepared dataset contains two orders of magnitude less subjects than cutting edge GAN-based 3D facial reconstruction methods, we show that it is possible to produce high-quality normalized face models for very challenging unconstrained input images, and demonstrate superior performance to the current state-of-the-art.

\end{abstract}
%%%%%%%%% BODY TEXT
\section{Introduction}

\begin{figure}[hbt!]
 \includegraphics[width=3.25in]{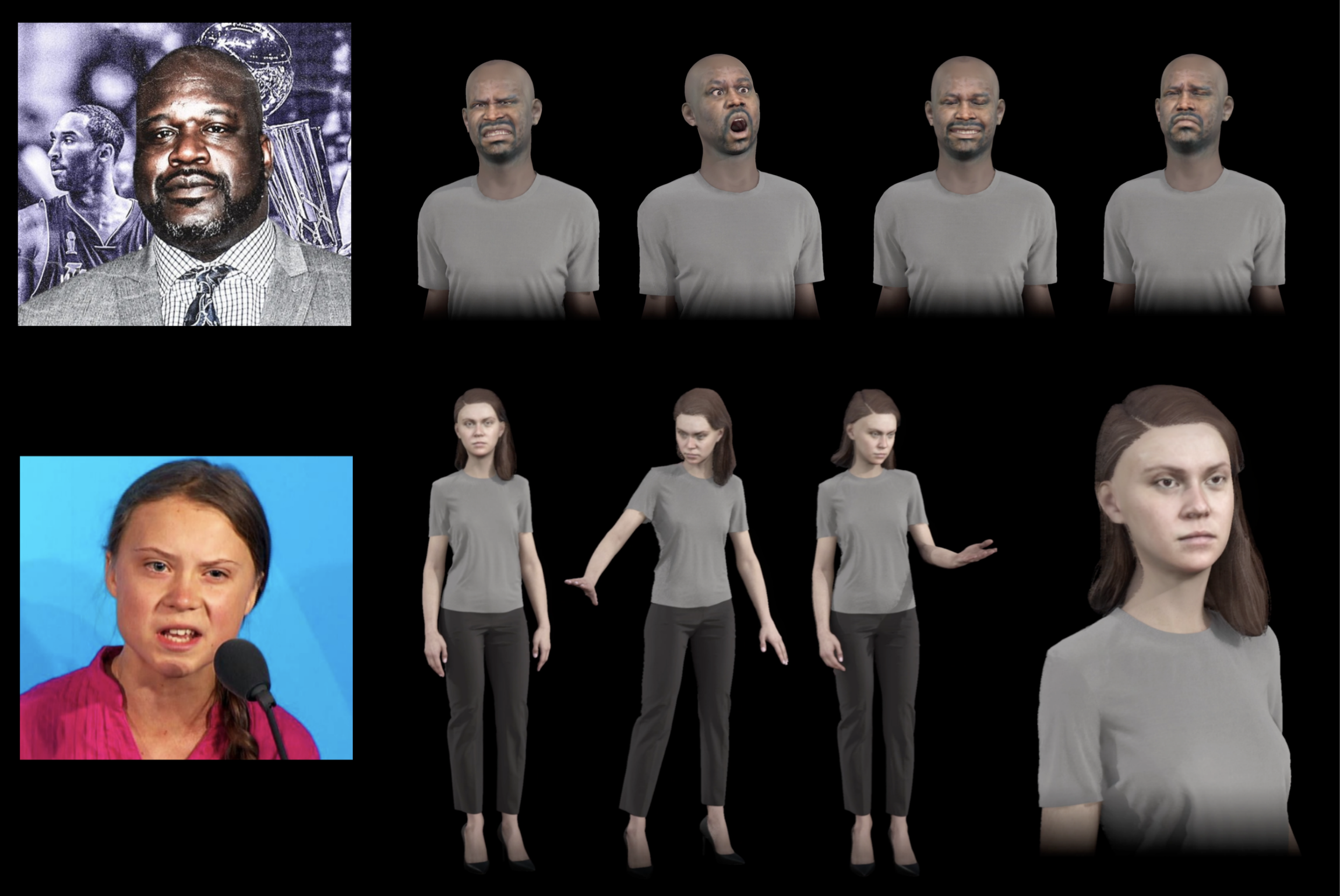}
 \caption{Automated digitization of normalized 3D avatars from a single photo.}
 \label{fig:rig}
\end{figure}

The creation of high-fidelity virtual avatars have been mostly reserved to professional production studios and typically involves sophisticated equipment and controlled capture environments. Automated 3D face digitization methods that are based on unconstrained images such as selfies or downloaded internet pictures are gaining popularity for a wide range of consumer applications, such as immersive telepresence, video games, or social media apps based on personalized avatars. %For instance, Spacial.io and Itseez3D uses 3D avatars to enable AR-based collaboration, Loom.ai turn users into avatars to make video conferencing more fun, and Pinscreen digitizes humans to create personalized virtual influencers and digital assistants. Consumer-accessible and personalized avatars have the potential to enable new forms of virtual connectivity and self-expression in a world where communication is becoming exclusively digital and remote.

Cutting-edge single-view avatar digitization solutions are based on non-linear 3D morphable face models (3DMM) generated from GANs~\cite{Tran_2018_CVPR,Tran_2019_CVPR,Gecer_2019_CVPR,Lee_2020_CVPR}, outperforming traditional linear models~\cite{Blanz_1999_SIGGRAPH} which often lack facial details and likeness of the subject. To successfully train these networks, hundreds of thousands of subjects in various lighting conditions, poses, and expressions are needed. While highly detailed 3D face models can be recovered, the generated textures have the lighting of the environment baked in, and expressions are often difficult to neutralize making these methods unsuitable for applications that require relighting or facial animation. In particular, inconsistent textured models are obtained when images are taken under different lighting conditions.

Collecting the same volume of 3D face data with neutral expressions and controlled lighting condition is intractable. Hence, we introduce a GAN-based facial digitization framework that can generate a high-quality textured 3D face model with neutral expression and normalized lighting using only thousands of real world subjects. Our approach consists of dividing the problem into two stages. The first stage uses a non-linear morphable face model embedded into a StyleGAN2~\cite{Karras_2020_CVPR} network to robustly generate detailed and clean assets of a normalized face. The likeness of the person is then transferred from the input photograph using a perceptual refinement stage based on iterative optimization using a differentiable renderer. StyleGAN2 has proven to be highly expressive in generating and representing real world images using an inversion step to convert image to latent vector~\cite{Abdal_2019_ICCV,Shen_2020_CVPR,Abdal_2020_CVPR,guan2020collaborative} and we are adopting the same two step GAN-inversion approach to learn facial geometry and texture jointly. To enable 3D neutral face inference from an input image, we connect the image with the embedding space of our non-linear 3DMM using an identity regression network based on identity features from FaceNet~\cite{Schroff_2015_CVPR}.
To train a sufficiently effective generator, we introduce a new \textit{Normalized Face Dataset} which consists of a combination of high-fidelity photogrammetry scans, frontal and neutral portraits in diffuse lighting conditions, as well as fake subjects generated using a pre-trained StyleGAN2 network with FFHQ dataset~\cite{Karras_2019_CVPR}. 

Despite our data augmentation effort, we show that our two-stage approach is still necessary to handle the large variation of possible facial appearances, expressions and lighting conditions. We demonstrate the robustness of our digitization framework on a wide range of extremely challenging examples, and provide extensive evaluations and comparisons with current state-of-the-art methods. Our method outperforms existing techniques in terms of digitizing textured 3D face models with neutral expressions and diffuse lighting conditions. Our normalized 3D avatars can be converted into parametric models with complete bodies and hair, and the solution is suitable for animation, relighting, and integration with game engines as shown in Fig.~\ref{fig:rig}.
We summarize our key contributions as follows:

\begin{itemize}
\item \vspace{-0.05in}We propose the first StyleGAN2-based approach for digitizing a 3D face model with neutral expressions and diffusely lit textures from an unconstrained image.  
\item \vspace{-0.1in}We present a two-stage digitization framework which consists of a robust normalized face model inference stage followed by a perception-based iterative face refinement step.
\item \vspace{-0.1in}We introduce a new data generation approach and dataset based on a combination of photogrammetry scans, photographs of expression and lighting normalized subjects, and generated fake subjects.
\item \vspace{-0.1in}Our method outperforms existing single-view 3D face reconstruction techniques for generating normalized faces, and we also show that our digitization approach works using limited subjects for training.
\end{itemize}
\section{Related Works}
\label{sec:relatedworks}

While a wide range of avatar digitization solutions exist for professional production, they mostly rely on sophisticated 3d scanning equipment (e.g., multi-view stereo, photometric stereo, depth sensors etc.) and controlled capture settings~\cite{Beeler_2010_SIGGRAPH, ghosh2011multiview, Fyffe_2016_EG}. We focus our discussion on monocular 3D face reconstruction methods as they provide the most accessible and flexible way of creating avatars for end-users, where only a selfie or downloaded internet photo is needed.

\paragraph{3D Morphable Face Models.}

Linear 3D Morphable Models (3DMM) have been introduced by Blanz and Vetter~\cite{Blanz_1999_SIGGRAPH} two decades ago, and have been established as the de-facto standard for 3D face reconstruction from unconstrained input images. The linear parametric face model encodes shape and textures using principal component analysis (PCA) built from $200$ laser scans. Various extensions of this work include the use of larger numbers of high-fidelity 3D face scans~\cite{Booth_2016_CVPR,Booth_2017_CVPR}, web images~\cite{Ira_2013_ICCV}, as well as facial expressions often based on PCA or Facial Action Coding Systems(FACS)-based blendshapes ~\cite{blanz2003reanimating,Vlasic:2005:FTM,cao2014facewarehouse}.

The low dimensionality and effectiveness of 3DMMs make them suitable for robust 3D face modeling as well as facial performance capture in monocular settings. To reconstruct a textured 3D face model from a photograph, conventional methods iteratively optimize for shape, texture, and lighting condition by minimizing energy terms based on constraints such as facial landmarks, pixel colors~\cite{Blanz_1999_SIGGRAPH,Sami_2005_CVPR,GVWT13,Shi:2014:AAH,Cao:2014:DDE,Ichim:2015:DAC,Thies_2016_CVPR,Garrido_2016_SIGGRAPH,cao2016real,Li_2017_SIGGRAPHASIA}, or depth information if available such as for the case of RGB-D sensors~\cite{weise09face,weise2011realtime,Bouaziz:2013:OMR,li2013realtime,hsieh2015unconstrained,hifi3dface2020tencentailab,Hu_2017_SIGGRAPHASIA}.

While robust face reconstruction is possible, linear face models combined with gradient optimization-based optimization are ineffective in handling the wide variation of facial appearances and challenging input photographs. For instance, detailed facial hair and wrinkles are hard to generate and the likeness of the original subject is typically lost after the reconstruction.
Deep learning-based inference techniques~\cite{wu2019mvf, Gecer_2019_CVPR,deng2019accurate,Genova_2018_CVPR,Tewari_2017_ICCV,Tran_2017_CVPR,Dou_2017_CVPR,Bas_2017_ICCV_Workshops,Tewari_2017_ICCV} were later introduced and have demonstrated significantly more robust facial digitization capabilities but they are still ineffective in capturing facial geometric and appearance detail due to the linearity and low dimensionality of the face model. Several post-processing techniques exist and use inferred linear face models to generate high-fidelity facial assets such as albedo, normal, and specular maps for relightable avatar rendering~\cite{Lattas_2020_CVPR,chen2019photo,Yamaguchi_2018}. AvatarMe~\cite{Lattas_2020_CVPR} for instance uses GANFIT~\cite{Gecer_2019_CVPR} to generate a linear 3DMM model as input to their post processing framework. Our proposed method can be used as alternative input to AvatarMe, and we compare it to GANFIT later in Section~\ref{sec:results}.

More recently, non-linear 3DMMs have been introduced. Instead of representing facial shapes and appearances as a linear combination of basis vectors, these models are formulated implicitly as decoders using neural networks where the 3D faces are generated directly from latent vectors. 
Some of these methods use fully connected layers or 2D convolutions in image space~\cite{Tran_2018_CVPR,Bagautdinov_2018_CVPR,feng2018prn,Tran_2019_CVPR,li2020learning}, while others use decoders in the mesh domain to represent local geometries~\cite{Litany_2018_CVPR, Ranjan_2018_ECCV, Zhou_2019_CVPR, Cheng_2019_MeshGAN, Abrevaya_2019_CVPR,Lee_2020_CVPR,Lin_2020_CVPR}. With the help of differentiable renderers~\cite{Tewari_2017_ICCV,Genova_2018_CVPR,ravi2020pytorch3d}, several methods~\cite{Tran_2018_CVPR,Tran_2019_CVPR,Lee_2020_CVPR} have demonstrated high-fidelity 3D face reconstructions using non-linear morphable face models using fully unsupervised or weakly supervised learning, which is possible using massive amounts of images in the wild. While the reconstructed faces are highly detailed and accurate w.r.t. the original input image, the generated assets are not suitable for relightable avatars nor animation friendly, since lighting conditions of the environment and expressions are baked into the output. 
Our work focuses on producing normalized 3D avatars with unshaded albedo textures and neutral expressions. Due to the limited availability of training data with normalized faces and the wide variation of facial appearances and capture conditions, the problem is significantly more challenging and ill-posed.

\paragraph{Generative Adversarial Network.}
We adopt StyleGAN2~\cite{Karras_2020_CVPR} to encode our non-linear morphable 3D face model. Among all generative models in deep learning, Generative Adversarial Networks (GANs)~\cite{NIPS2014_5423} have achieved a great success in producing realistic 2D natural images, nearly indistinguishable from real world images. After a series of advancements, state-of-the-art GANs like PGGAN~\cite{karras2018progressive}, BigGAN~\cite{brock2018large} and StyleGAN/StyleGAN2~\cite{Karras_2019_CVPR,Karras_2020_CVPR} have proven to be also effective in generating high resolution images and the ability to handle an extremely wide range of variations. In this work, we mainly focus on adopting StyleGAN2~\cite{Karras_2020_CVPR} to jointly learn facial geometry and texture, 
since its intermediate latent representation has been proven effective %to encode layer-wise style $\textbf{w}$ (i.e. independent embedding for each layer)
to best reconstruct a plausible target image with clean assets~\cite{Abdal_2019_ICCV,Shen_2020_CVPR,Abdal_2020_CVPR,guan2020collaborative}.

\paragraph{Facial Image Normalization.}
To address the problem of unwanted lighting and expressions during facial digitization, several methods have been introduced to normalize unconstrained portraits. Cole et al.~\cite{cole2017synthesizing} introduced a deep learning-based image synthesis framework based on FaceNet's latent code~\cite{Schroff_2015_CVPR}, allowing one to generate a frontal face with neutral expression and normalized lighting from an input photograph. More recently, Nagano et al~\cite{Nagano_2019_Siggraph} improved the method to generate higher resolution facial assets for the purpose generating high-fidelity avatars. In particular, their method breaks down the inference problem into multiple steps, solving explicitly for perspective undistortion, lighting normalization, followed by pose frontalization and expression neutralization. While the successful normalized portraits were demonstrated, their method rely on transferring details from the input subject to the generated output.
Furthermore, both methods rely on the linear 3DMMs for expression neutralization and thus cannot capture detailed appearance variations.
Neutralizing expression from nonlinear 3DMM, however, is not straightforward since the feature space of identity and expression are often entangled. Our new normalization framework with GAN-based reconstruction fills in this gap.

\section{Normalized 3D Avatar Digitization}
\label{sec:methods}

\begin{figure}[hbt!]
 \includegraphics[width=3.2in]{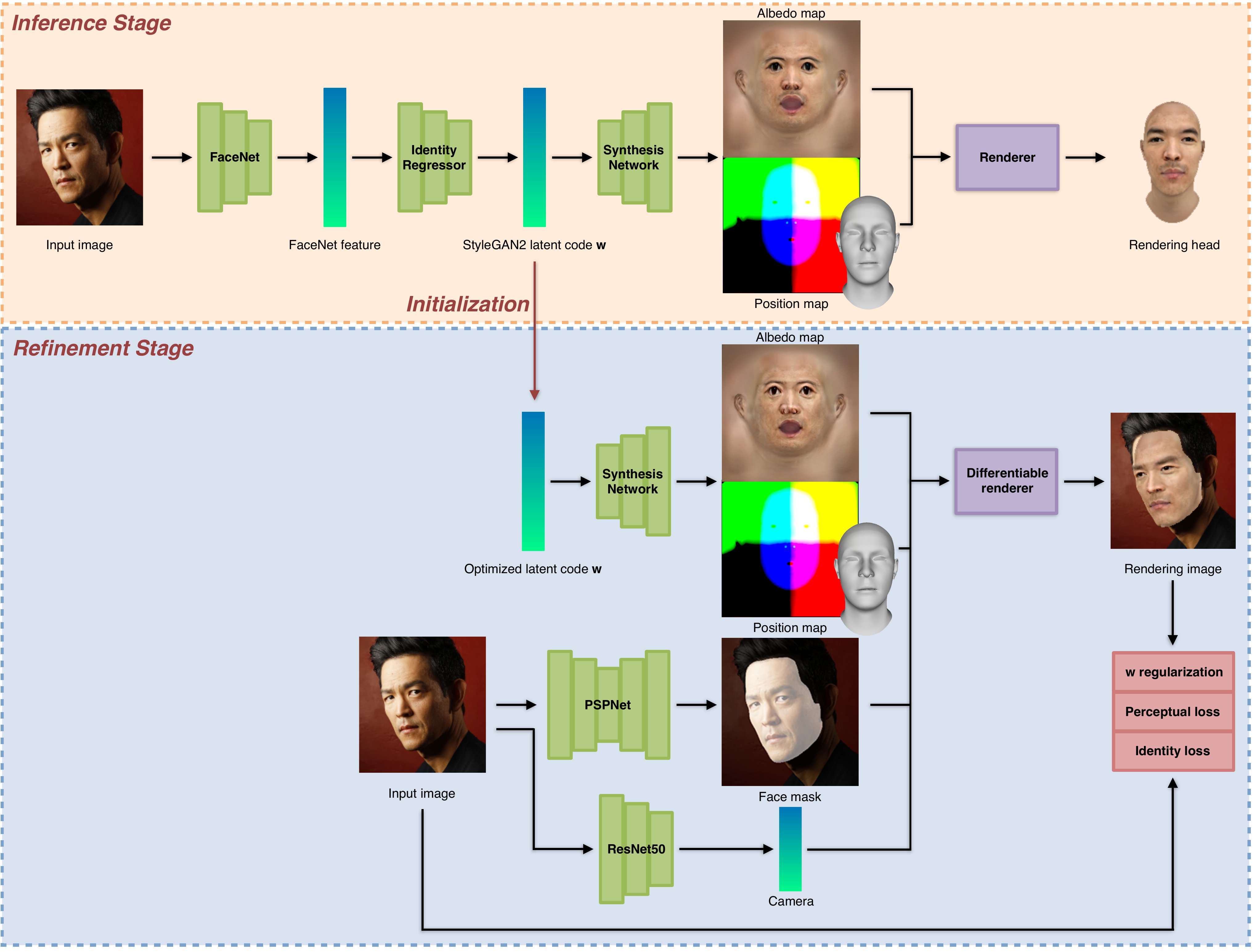}
 \caption{Two-stage facial digitization framework. The avatar is firstly predicted in the inference stage, and then improved to match the input image in the refinement stage.}
 \label{fig:inference_overview}
\end{figure}

An overview of our two-stage facial digitization framework is illustrated in Fig.~\ref{fig:inference_overview}. At the inference stage, our system uses a pre-trained face recognition network FaceNet~\cite{Schroff_2015_CVPR} to extract a person-specific facial embedding feature given an unconstrained input image. This identity feature is then mapped to the latent vector $\textbf{w} \in \mathcal{W}+$ in the latent space of our \textit{Synthesis Network} using an \textit{Identity Regressor}. The synthesis network decodes $\textbf{w}$ to an expression neutral face geometry and a normalized albedo texture. For the refinement, the latent vector $\textbf{w}$ produced by the inference is then optimized iteratively using a differentiable renderer by minimizing the perceptual difference between the input image and the rendered one via gradient descent.

\subsection{Robust GAN-Based Facial Inference}
\label{sec:inference_pipeline}

Our synthesis network $G$ generates the geometry as well as the texture in UV space. Each pixel in the UV map represents the 3D position and the RGB albedo color of the corresponding vertex using a 6-channel tuple $(\textsf{r},\textsf{g},\textsf{b},\textsf{x},\textsf{y},\textsf{z})$. The synthesis network is first trained using a GAN to ensure robust and high quality mapping from any normal distributed latent vector $\mathcal{Z}\sim \mathcal{N}(\mu ,\sigma)$. Then, the identity regression network $R$ is trained by freezing $G$ to ensure accurate mapping from the identity feature of an input image. Further details of each network are described below.

\begin{figure}[hbt!]
 \includegraphics[width=3.2in]{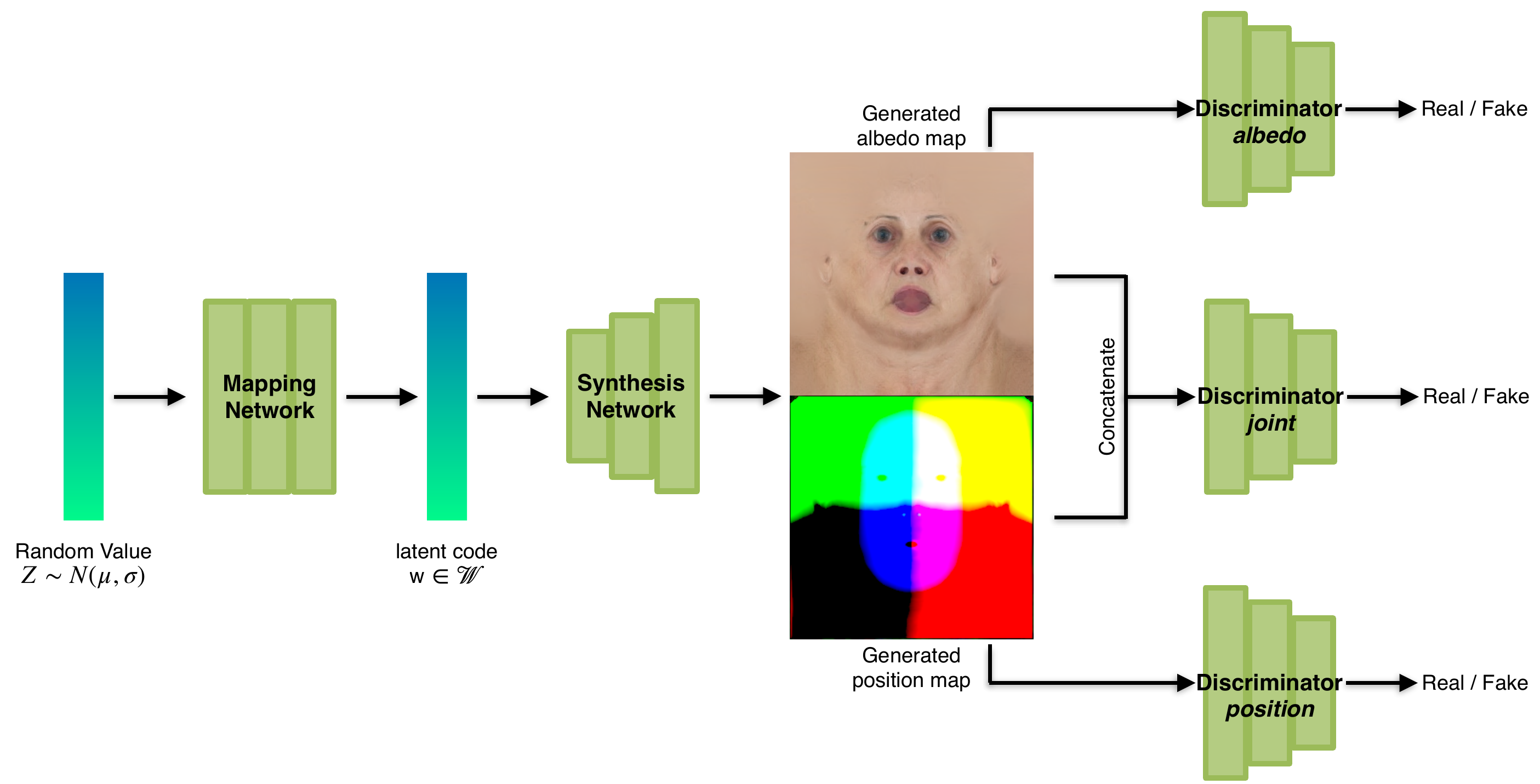}
 \caption{GAN-based geometry and texture synthesis.}
 \label{fig:generator}
\end{figure}

We train our synthesis network to embed a nonlinear 3D Morphable Model into its latent space, in order to model the cross correlation between the 3D neutral face geometry and the neutral albedo texture, as well as to generate high fidelity and diverse 3D neutral faces from a latent vector. Inspired by~\cite{li2020learning}, we adopt the StyleGAN2~\cite{Karras_2020_CVPR} architecture to train a morphable face model using 3D geometry and albedo texture as shown in Fig.~\ref{fig:generator}. {Rather than predicting vertex positions directly, we infer vertex position offsets relative to the mean face mesh to improve numerical stability.} To jointly learn geometry and texture, we project the geometry representation of classical linear 3DMMs $S \in \mathbb{R}^{3 \times N} $, which consists of a set of $N = 13557$ vertices on the face surface, onto a UV space using cylindrical parameterization. The vertex map is then rasterized to a 3-channel position map with $256 \times 256$ pixels. Furthermore, we train 3 discriminators jointly, including 2 individual ones for albedo and vertex position as well as a joint discriminator taking both maps as input. The individual discriminators ensure the quality and sharpness of each generated map, while the joint discriminator can learn and preserve their correlated distribution. This GAN is trained solely from the provided ground truth 3D geometries and albedo textures without any knowledge of the identity features.

\begin{figure}[hbt!]
 \includegraphics[width=3.2in]{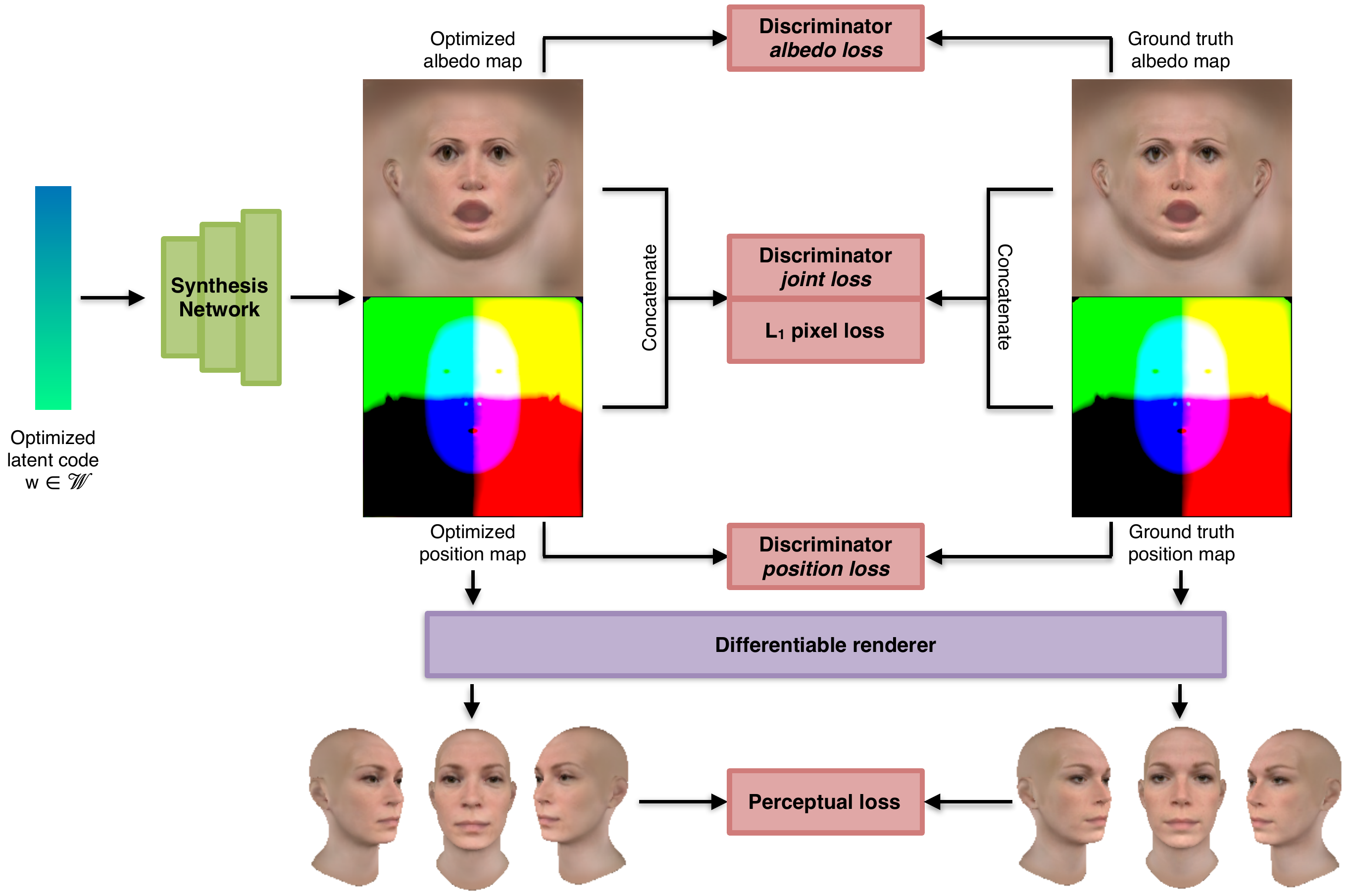}
 \caption{Our GAN-inversion searches a corresponding $\textbf{w}$, which can reconstruct the target geometry and texture.}
 \label{fig:projection}
\end{figure}

After obtaining $G$, we retrieve the corresponding input latent code via our code inversion algorithm. Inspired by~\cite{Abdal_2019_ICCV,zhu2019lia}, we choose the disentangled and extended latent space $\mathcal{W}+ := \mathbb{R}^{14 \times 512}$ of StyleGAN2 as the inversion space to achieve better reconstruction accuracy. As shown in Fig.~\ref{fig:projection}, we adopt an optimization approach to find the embedding of a target pair of position and albedo map with the following loss function:
\begin{equation}
    L_{inv} = L_{pix} + \lambda_{1}L_{\text{LPIPS}} + \lambda_{2}L_{adv}
\label{eq:L_inversion}
\end{equation}
where $L_{pix}$ is the $L_1$ pixel error of the synthesized position and texture maps, $L_{\text{LPIPS}}$ is the LPIPS distance~\cite{zhang2018perceptual} as a perceptual loss, and $L_{adv}$ is the adversarial loss favoring realistic reconstruction results using the three discriminators trained with $G$. Note that while LPIPS outperforms other perceptual metrics in practice~\cite{zhang2018perceptual}, it is trained with real images and measuring the perceptual loss directly on our UV maps would lead to unstable results. Therefore, we use a differentiable renderer~\cite{ravi2020pytorch3d} to render the geometry and texture maps from three fixed camera viewpoints and compute the perceptual loss based on these renderings.
Finally, the identity regressor $R$ can be trained using the solved latent codes of the synthesis network and their corresponding identity features from the input images.

\subsection{Unsupervised Dataset Expansion}
\label{sec:frontal_neutral_dataset}

\begin{figure}[hbt!]
 \includegraphics[width=3.2in]{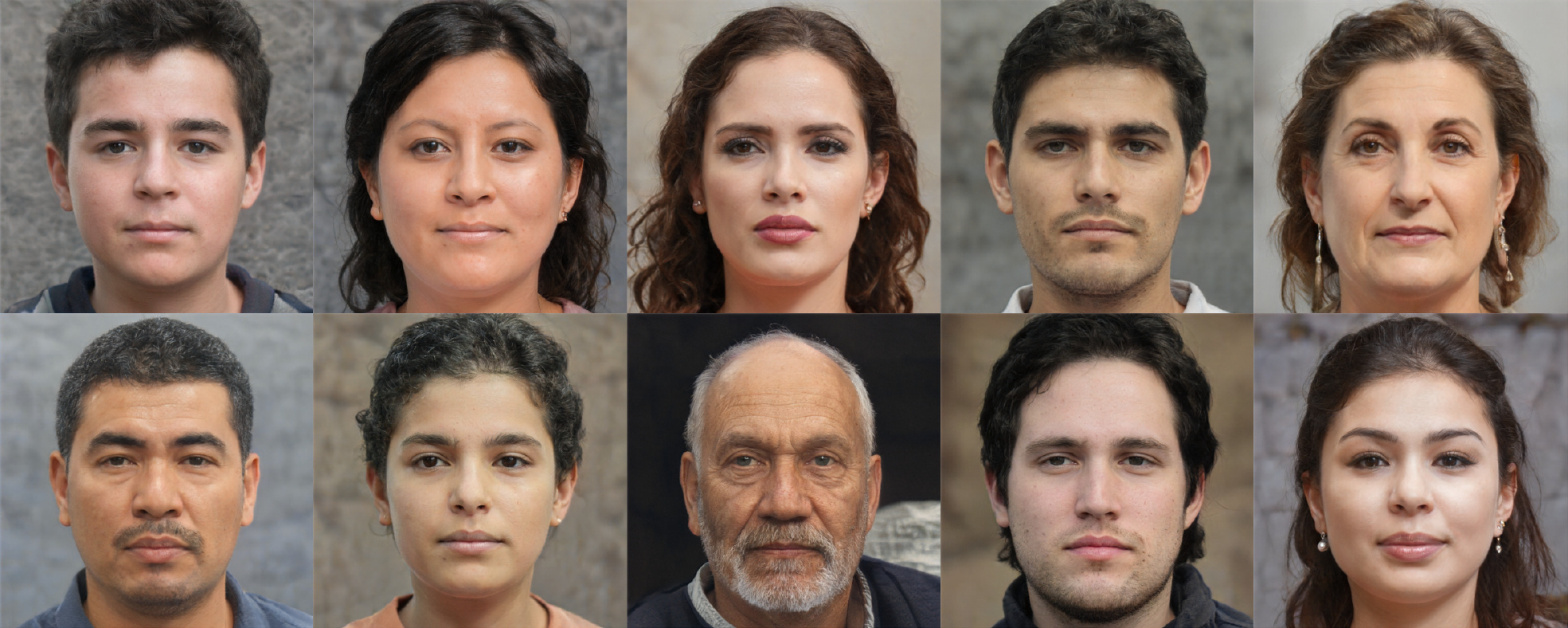}
 \caption{Examples of synthetic faces from our \textit{Normalized Face Dataset}.}
 \label{fig:fake_person}
\end{figure}

While datasets exist for frontal human face images in neutral expression~\cite{Ma_2015_data, du2014compound, GROSS2010807, doi:10.1080/02699930903485076}, the amount of such data is still limited and the lighting conditions often vary between datasets. Instead of manually collecting more images from the Internet for expanding our training data, we propose an automatic approach to produce frontal neutral portraits based on the pre-trained StyleGAN2 network trained with FFHQ dataset. {Similar to a recent technique for semantic face editing~\cite{Shen_2020_CVPR}, we train a neural network to predict identity attributes $\alpha$ of an input image in latent space. We used images collected from internet as input and estimate each $\alpha$ and apply it to $\textbf{w}_{mean}$. $\textbf{w}_{mean}$ is a fixed value in latent space, which could generate a mean and frontalized face. We then use a latent editing vector $\beta$ to neutralize the expressions. The final latent value $\textbf{w}' = \textbf{w}_{mean} + \alpha + \beta$ produces a frontalized and neutralized face by feeding into StyleGAN2.}
%We further run off-the-shelf head pose and expression networks\footnote{APIs provided by popular cloud service providers e.g. Google Cloud, Azure, and AWS are evaluated and lead to consistent results.} to verify that these synthetic faces are indeed frontal and neutral.
Some examples are shown in Fig.~\ref{fig:fake_person}. We further emphasize that all images in our \textit{Normalized Face Dataset} are frontal and have neutral expressions. Also, these images have well conditioned diffuse scene illuminations, which are preferred for conventional gradient descent-based 3D face reconstruction methods.

For each synthesized image, we apply light normalization~\cite{Nagano_2019_Siggraph} and 3D face fitting based on Face2Face~\cite{Thies_2016_CVPR} to generate a 3D face geometry and then project the light normalized image for the albedo texture. Instead of using the linear 3DMM completely, which results in coarse and smooth geometry, we first run our inference pipeline to generate the 3D geometry and take it as the initialization for the Face2Face optimization. After optimization, the resulting geometry is in fact the non-linear geometry predicted from our inference pipeline plus a linear combination of blendshape basis optimized by Face2Face, thus preserving its non-linear expressiveness. Also note that the frontal poses of the input images facilitate our direct projections onto UV space to reconstruct high-fidelity texture maps.

The complete training procedure works as follows: we first collect a high quality \textit{Scan Dataset} with $431$ subjects with accurate photogrammetry scans, with $63$ subjects from 3D Scan Store~\cite{3D_scan_store} and $368$ subjects from Triplegangers~\cite{tripelgangers}. The synthesis network $G_0$ is then trained from such scan data, and is then temporarily frozen for latent code inversion and the training of identity regressor $R_0$. These bootstrapping networks $(R_0, G_0)$ trained on the small \textit{Scan Dataset} are applied onto our \textit{Normalized Face Dataset} to infer the geometry and texture, which are then optimized and/or corrected by the Face2Face algorithm. Next, the improved geometry and texture are added back into the training of $(R_0, G_0)$ to obtain the fine-tuned networks $(R_1, G_1)$ with improved accuracy and robustness. 

%The new predictions can still be further improved by Face2Face until network training and geometry optimization both converge. More images can then be added into the \textit{Normalized Face Dataset} for the next round of training and optimization.
Our final \textit{Normalized Face Dataset} consists of $5601$ subjects, with $368$ subjects from Triplegangers, $597$ from Chicago Face Dataset (CFD)~\cite{Ma_2015_data}, $230$ from the compound facial expressions (CFE) dataset~\cite{du2014compound}, $153$ from The CMU Multi-PIE Face Dataset~\cite{GROSS2010807}, $67$ from Radboud Faces Database (RaFD)~\cite{doi:10.1080/02699930903485076}, and the remaining $4186$ generated by our method. We use most of the frontal and neutral face images that are available to increase diversity, but still rely on the large volume of synthetic data for the training.

\begin{figure*}[hbt!]
\centering
 \includegraphics[width=6.5in]{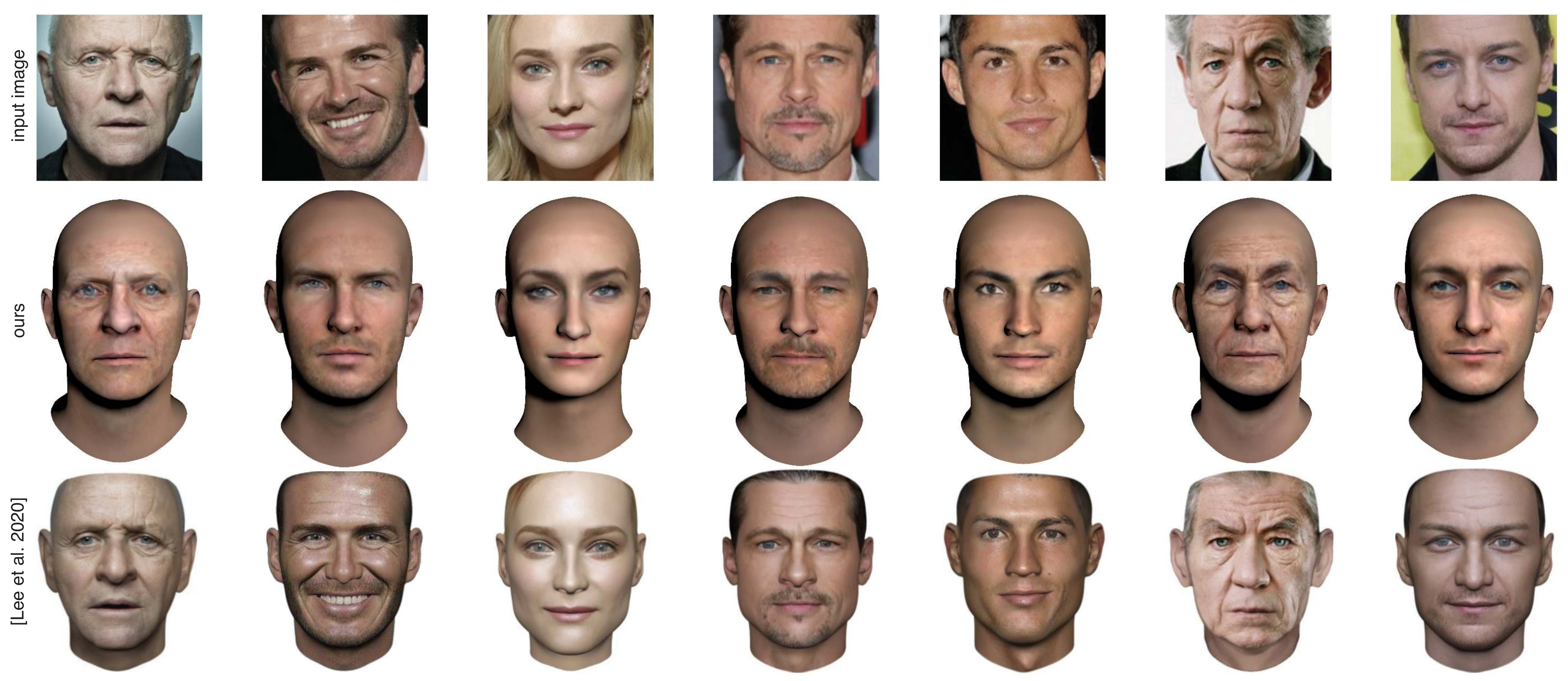}
 \caption{Qualitative comparison with other state-of-the-art 3D face reconstruction method. The first row shows the input images and the second row shows our results, and the third row are the reconstructed 3D faces obtained by \cite{Lee_2020_CVPR}.}
 \label{fig:comparison}
\end{figure*}

\subsection{Perceptual Refinement}
\label{sec:perceptual_optimization}
While the inference pipeline described in Sec.~\ref{sec:inference_pipeline} with training data from Sec.~\ref{sec:frontal_neutral_dataset} can reliably infer the normalized texture and geometry from an unconstrained image, a second stage with perceptual refinement can help determine a neighbor of the predicted latent code in the embedding space that matches the input image better. The work from Shi et al.~\cite{shi2019probabilistic} shows that an embedding space learned for face recognition is often noisy and ambiguous due to the nature of fully unconstrained input data. While FaceNet predicts the most likely latent code, the variance (or \textit{uncertainty} in Shi et al.'s work) could be large. A small perturbation of the latent code may not affect the identity feature training at all. On the other hand, such a small error in the identity code may cause greater inconsistency in our inference pipeline after passing $R$ and $G$.

An ``end-to-end'' refinement step is introduced, to handle never seen before images while ensuring consistency between the final renderings using the predicted geometry and texture, and the input image. Fig.~\ref{fig:inference_overview} shows the end-to-end architecture for this refinement step. We reuse the differentiable renderer to generate a 2D face image $\hat{I}$ from the estimated 3D face, and compute the perceptual distance with the input image $I$. To project the 3D face back to the head pose in image $I$, we train a regression network with ResNet50~\cite{He_2016_CVPR} as backbone to estimate the camera $c=[t_x,t_y,t_z,r_x,r_y,r_z,f]^T$ from $I$, where $[t_x,t_y,t_z]^T$ and $[r_x,r_y,r_z]^T$ denote the camera translation and rotation and $f$ is the focal length. The network is trained using the accurate camera data from the \textit{Scan Dataset} and the estimated camera data from \textit{Normalized Face Dataset}, computed by Face2Face. Furthermore, in order to blend the projected face only image with the background from the original image $I$, we train a PSPNet~\cite{zhao2017pspnet} with ResNet101~\cite{He_2016_CVPR} as backbone using CelebAMask-HQ~\cite{CelebAMask-HQ}. We then blend the rendered image $\hat{I}$ into the segmented face region from $I$ {to produce $I_{0}$}. 
The final loss is simply represented as:
\begin{equation}
    L_{refine} = L_{w} + \lambda_1 L_{\text{LPIPS}} + \lambda_2 L_{id}\quad,
\label{eq:L_refine}
\end{equation}
where $L_{w}$ is a regularization term on $\textbf{w}$, i.e., the Euclidean distance between the variable $\textbf{w}$ and its initial prediction derived by $R$, enforcing the similarity between the modified latent and the initial prediction. $L_{LPIPS}$ is the perceptual loss measured by LPIPS distance~\cite{zhang2018perceptual} between $I_{0}$ and $I$, which enables improved matching in terms of robustness and better preservation of semantically meaningful facial features compared to using pixel differences. $L_{id}$ is the cosine similarity between the identity feature of $\hat{I}$ and $I$, to preserve consistent identity. 
\section{Results}
\label{sec:results}

We demonstrate the performance of our method in Fig.~\ref{fig:teaser} and~\ref{fig:comparison}, and show how our method can handle extremely challenging unconstrained photographs with very harsh illuminations, extreme filtering, and arbitrary expressions. We can produce plausible textured face models where the likeness of the input subject is preserved and visibly recognizable. Compared to the state-of-the-art 3D face reconstruction method (see Fig.~\ref{fig:comparison}) based on non-linear 3DMMs, our method can neutralize expressions and produce an unshaded albedo texture suitable for rendering in arbitrary lighting conditions as demonstrated using various HDRI-based lighting environments. We also show in Fig.~\ref{fig:rig} how we can obtain a fully rigged 3D avatar from a single photo including body and hair, by adopting the hair digitization algorithm in~\cite{Hu_2017_SIGGRAPHASIA} (see accompanying video for live demo). 

\paragraph{Evaluations.}

\begin{figure}[hbt!]
 \includegraphics[width=3.25in]{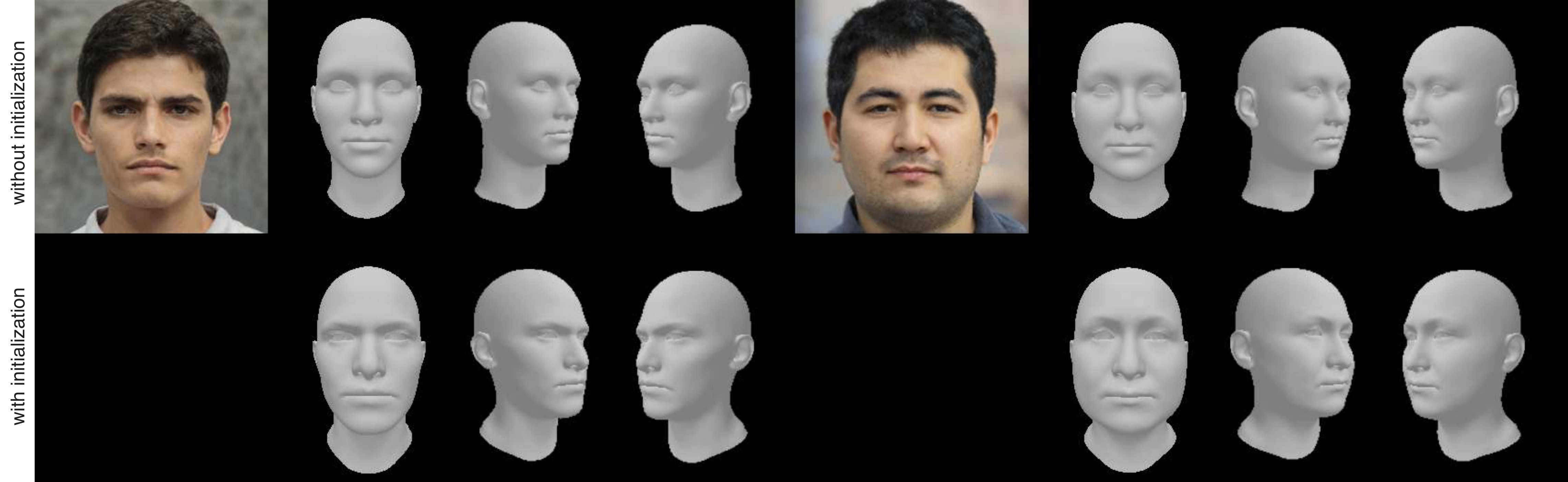}
 \caption{Face2Face optimization results. The first row is the original implementation~\cite{Thies_2016_CVPR}. The second row is our proposed improvement with nonlinear initialization.}
 \label{fig:f2f}
\end{figure}

Sec.~\ref{sec:frontal_neutral_dataset} further improves the performance of $G$ and $R$ using more training data. Fig.~\ref{fig:f2f} compares the default Face2Face optimization using a linear 3DMM with the improved ones using an initialization from $R_0$ and $G_0$. %The GAN-based initialization highlights the performance of the nonlinear property in the reconstruction results and makes the training for $R_1$ and $G_1$ possible with such synthetic but accurate and expressive data.

\begin{figure}[hbt!]
 \includegraphics[width=3.25in]{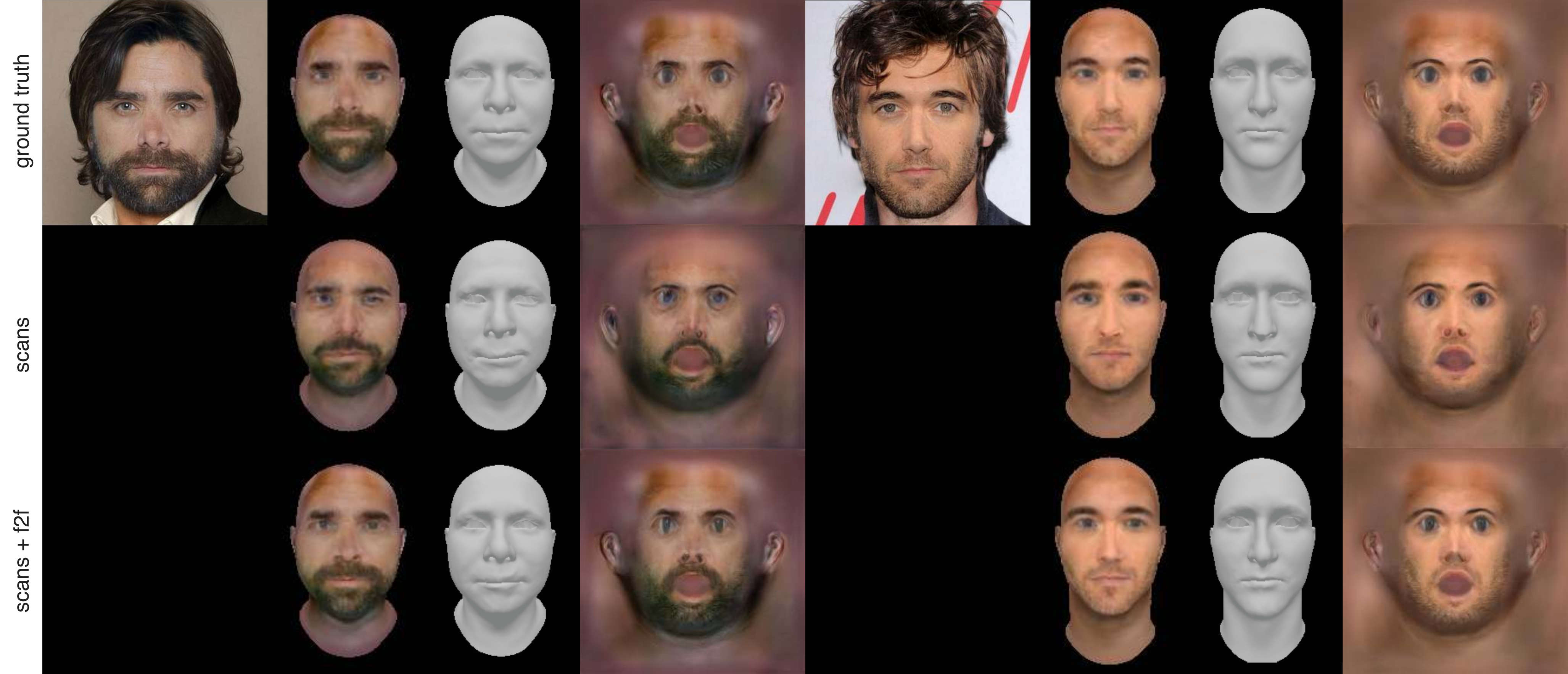}
 \caption{Expressiveness of the synthesis network trained with different datasets. From top to bottom: The ground truth; The GAN-inversion results based on $G_0$ trained with \textit{Scan Dataset} only; The same process based on $G_1$, trained with \textit{Normalized Face Dataset}.}
 \label{fig:projection_data}
\end{figure}

\begin{figure}[hbt!]
 \includegraphics[width=3.25in]{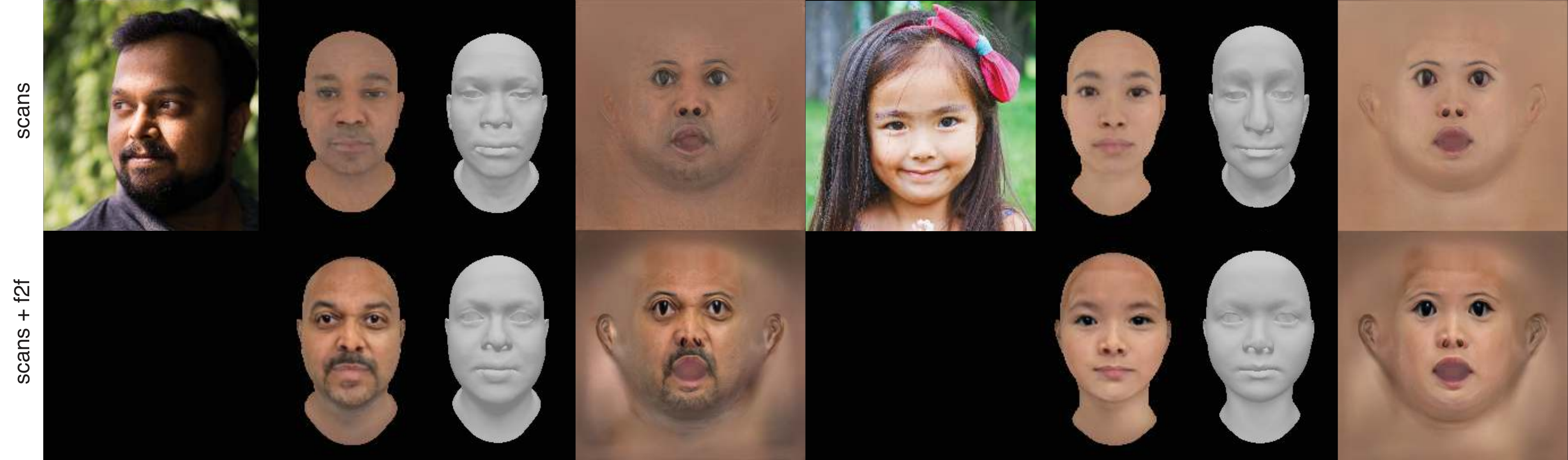}
 \caption{Quality of the regression network trained with different datasets. The first row shows the inference results by $R_0$m trained with \textit{Scan Dataset}. The second row shows the results by $R_1$, trained with \textit{Normalized Face Dataset}.}
 \label{fig:regression_data}
\end{figure}

\begin{figure}[hbt!]
 \includegraphics[width=3.25in]{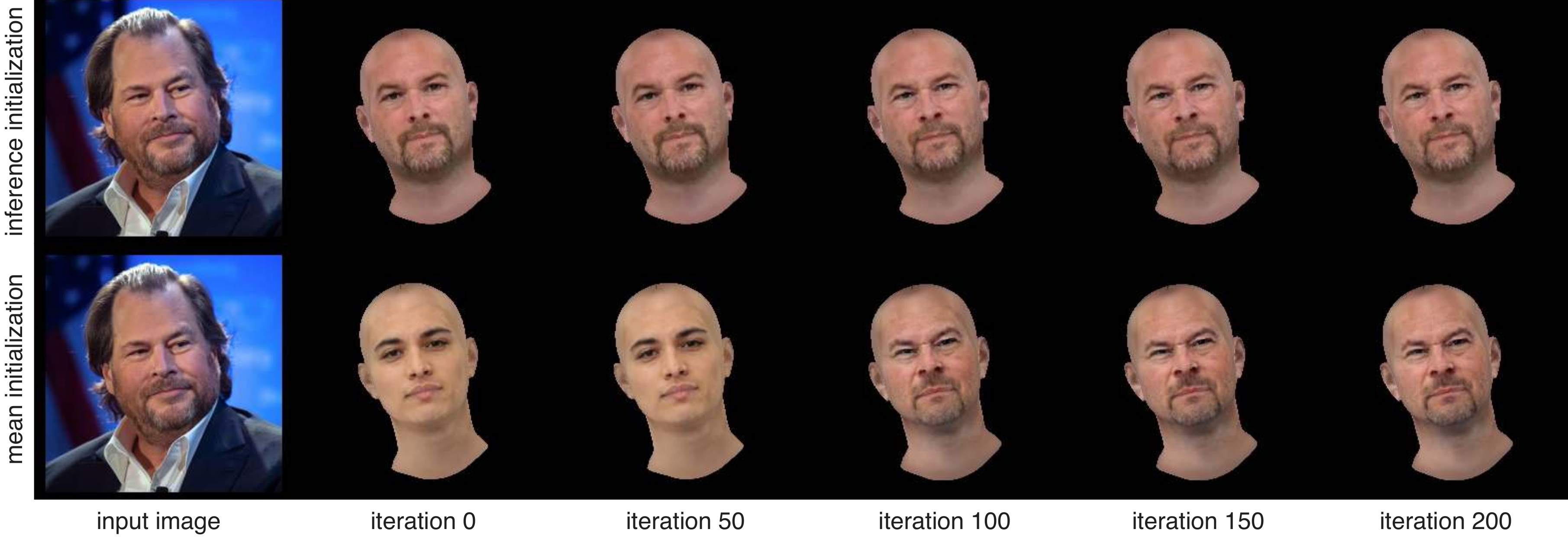}
 \caption{Qualitative comparison with different initialization schemes for iterative refinement. %We choose the refine-ment results at different iterations to show the trend of therefinement procedure.
 The mean initialization starts optimization from a mean latent vector of our training dataset. The inference initialization starts from the latent vector predicted by $R$.}
 \label{fig:optimization_initialization}
\end{figure}

With such synthetic training data, Fig.~\ref{fig:projection_data} shows improved expressiveness of $G_1$ than $G_0$. Several artifacts from $G_0$ around eyes and the lack of facial hair are fixed in $G_1$. In Fig.~\ref{fig:regression_data}, $R_1$ also shows higher diversity of face shapes and superior accuracy compared to $R_0$ after training with the \textit{Normalized Face Dataset}.
Fig.~\ref{fig:optimization_initialization} demonstrates the effect of both the inference stage in Sec.~\ref{sec:inference_pipeline} and the refinement stage. For each row of the experiment, the end-to-end iterative refinement can always improve the likeness and expressiveness of the 3D avatar. However, notice that the refinements from the mean latent vector would fail to produce a faithful result after $200$ iterations, while the refinements from an accurate initial prior by $R$ converges to a highly plausible face reconstruction.

\begin{figure}[hbt]
 \includegraphics[width=3.25in]{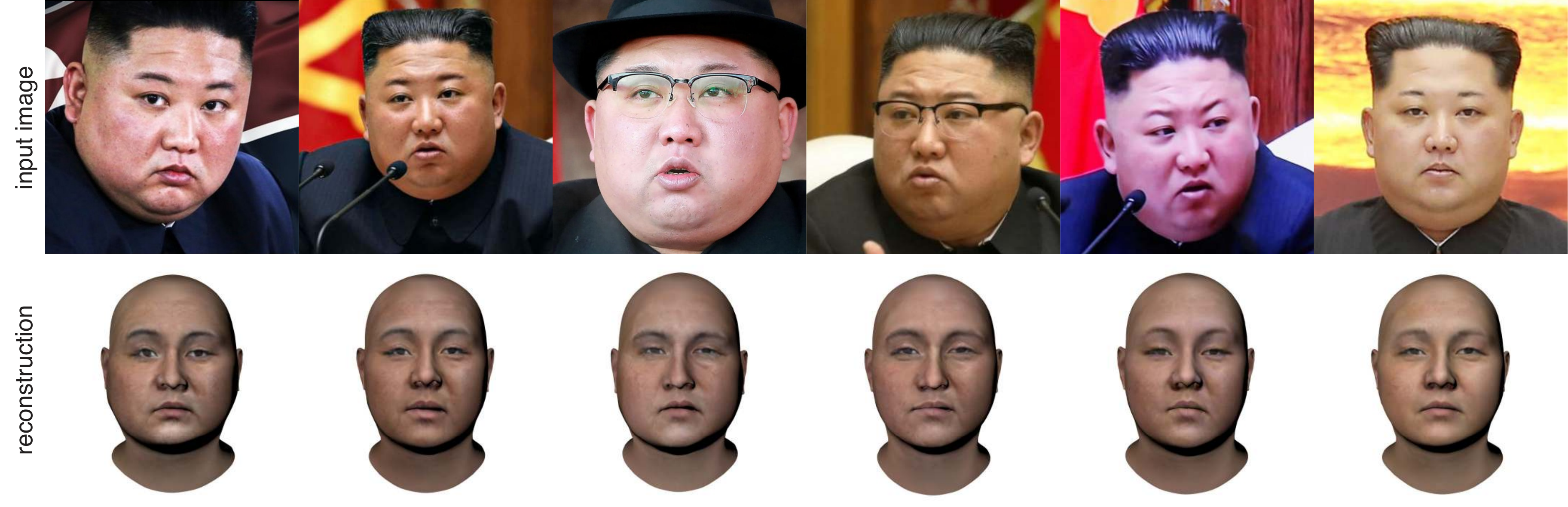}
 \caption{Consistent reconstructions of the same person under different environments.}
 \label{fig:expressions_consistency}
\end{figure}

Since our proposed pipeline simply rely on the identity and perceptual features from $I$, the reconstructed 3D avatar is invariant to the factors FaceNet filters, such as occlusion, image resolution, lighting environment, and facial expression.
Fig.~\ref{fig:expressions_consistency} demonstrates how we can obtain consistent geometries from different lighting, viewpoints, and facial expressions. Further results of more challenging images, such as low resolution or largely occluded ones are provided in the supplemental material.

\paragraph{Comparisons.}

Fig.~\ref{fig:comparison} compare our method with the most recent single view face reconstruction method~\cite{Lee_2020_CVPR}. Lee et al.~\cite{Lee_2020_CVPR} adopts a state-of-the-art nonlinear 3DMM on both geometry and texture. They also use a Graph Convolutional Neural Network to embed geometry and a Generative Adversarial Network to synthesize texture. However, they train two networks separately with different datasets, where facial shape and appearance are uncorrelated. More importantly, their results show that expressions and lighting are baked in, which makes their method unsuitable for relighting and facial animation purposes. More comparisons with other monocular face reconstruction methods~\cite{deng2019accurate, Gecer_2019_CVPR, Tran_2019_CVPR} can be found in the supplemental material.

\begin{figure}[hbt]
 \includegraphics[width=3.25in]{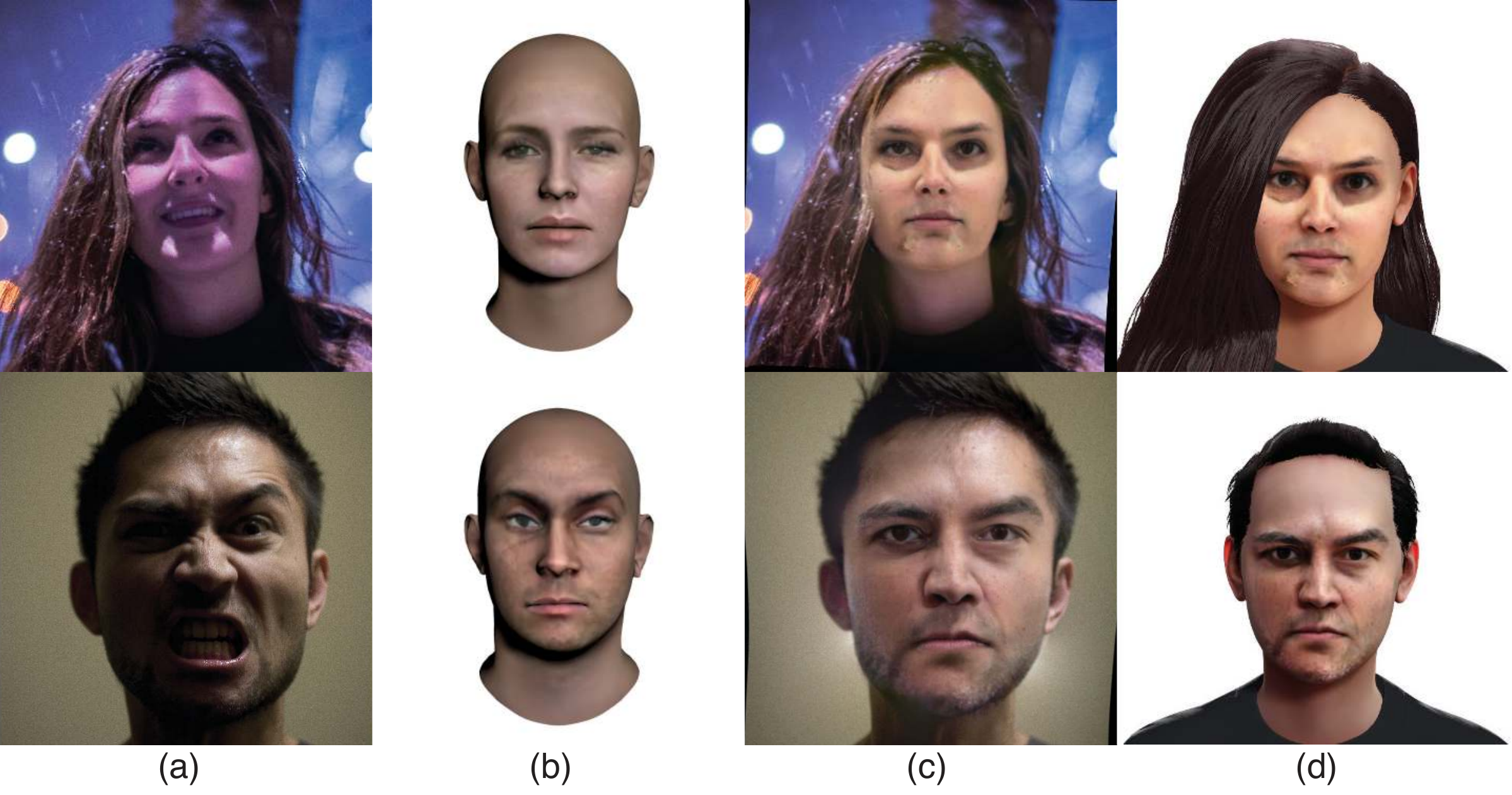}
 \caption{Qualitative comparison with state-of-the-art face normalization method~\cite{Nagano_2019_Siggraph}. From left to right, we show (a) input image; (b) our reconstructed result; (c) image-based face normalization result generated by Nagano et al.~\cite{Nagano_2019_Siggraph}; (d) Face2Face reconstruction result based on (c).}
 \label{fig:koki}
\end{figure}

Fig.~\ref{fig:koki} shows our results compared to the deep face normalization method~\cite{Nagano_2019_Siggraph}. While some successful normalized results were demonstrated, their image-to-image translation architecture transfers details from the input subject to the generated output. If those details are deteriorated, then face normalization would fail.

\begin{table}[h]
\begin{center}
\begin{tabular}{c|c|c}
\hline
 Tran et al.~\cite{Tran_2019_CVPR} & Deng et al.~\cite{deng2019accurate} & Ours \\ 
 \hline
 1.935mm & 1.568mm & \textbf{1.557mm} \\
 % 1.93459mm & 1.56805mm & \textbf{1.55741mm} \\
 \hline
\end{tabular}
\caption{Quantitative comparison of with other 3D face reconstruction methods.}
\label{table:facescape_comparison}
\end{center}
\vspace{-0.1in}
\end{table}

\begin{table}[h]
\begin{center}
\begin{tabular}{c|c|c}
\hline
 Tran et al.~\cite{Tran_2019_CVPR} & Deng et al.~\cite{deng2019accurate}& Ours \\ 
 \hline
 0.304 & 0.392 & \textbf{0.205} \\
 \hline
\end{tabular}
\caption{Quantitative comparison on texture.}
\label{table:texture_comparison}
\end{center}
\vspace{-0.1in}
\end{table}

%\textcolor{red}{quantitative comparison}
Quantitative experiments on FaceScape~\cite{yang2020facescape} using high resolution 3D scans and corresponding images are shown in Tables~\ref{table:facescape_comparison} and ~\ref{table:texture_comparison}. For geometric accuracy, we randomly select $20$ scans from FaceScape, and for each method, we compute the average point to mesh distance between the monocular reconstructed geometry and the ground truth scan. The proposed model has smaller reconstruction errors than other state-of-the-art ones. For texture evaluation, we augment the input images with lighting variations and compute the mean L1 pixel loss between generated textures from each method and the ground truth. Our method generates textures that are less sensitive to lighting conditions. 

\paragraph{Implementation Details.}

All our networks are trained on a desktop machine with Intel i7-6800K CPU, 32GB RAM and one NVIDIA TITAN GTX (24GB RAM) GPU using PyTorch~\cite{paszke2019pytorch}. The StyleGAN2 network training takes $13$ days with the \textit{Normalized Face Dataset}. We use the PyTorch implementation~\cite{stylegan2-pytorch} and remove the noise injection layer in the original implementation to remove the stochastic noise inputs and enable full control of the generated results from the latent vector. Our identity regression network is composed of four fully connected layers with Leaky ReLU activations, and the training takes $1$ hour to converge with the same training data. 
At the testing stage, inference takes $0.13$ s and refinement takes $45$ s for $200$ iterations.

\begin{figure}[hbt!]
 \includegraphics[width=3.25in]{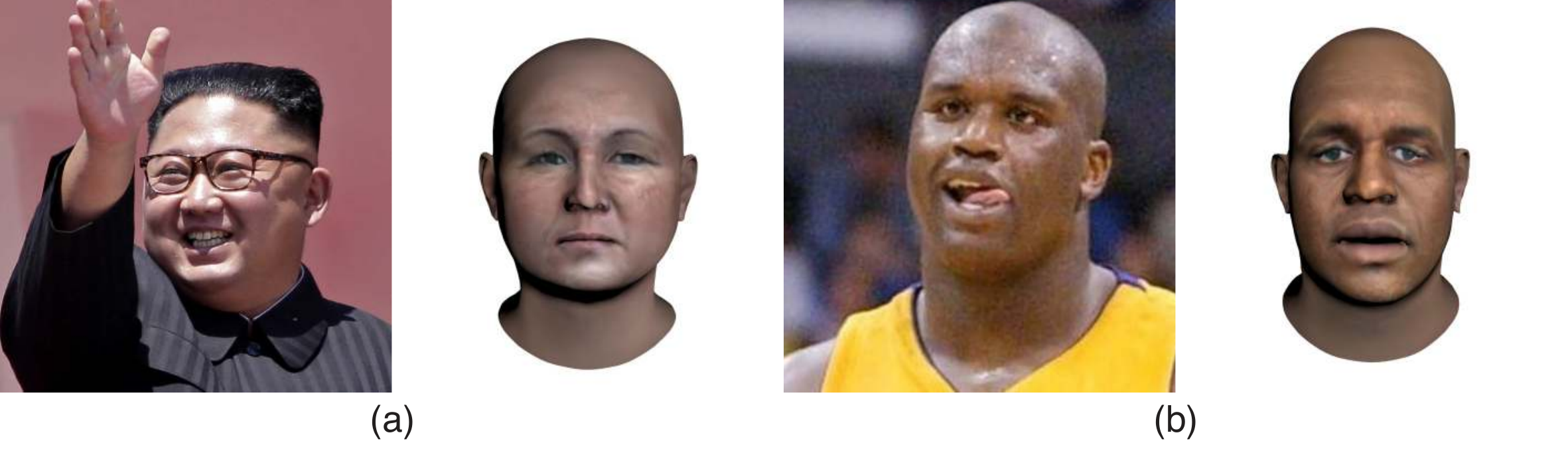}
 \caption{Failure cases in our experiments. (a) shows a failure where the specularity at the chin is baked into the generated result; (b) shows that the robustness of the reconstruction result is affected by the exaggerated expression.}
 \label{fig:failure_case}
\end{figure}

\section{Discussion}

We have demonstrated a StyleGAN2-based digitization approach using a non-linear 3DMM that can reliably generate high-quality normalized textured 3D face models from challenging unconstrained input photos. Despite the limited amount of available training data (only thousands of subjects), we have shown that our two-stage face inference method combined with a hybrid \textit{Normalized Face Dataset} is effective in digitizing relightable and animation friendly avatars and can produce results of quality comparable to state-of-the-art techniques where generated faces are not normalized.
Our experiments show that simply adopting existing methods using limited normalized facial training data is insufficient to capture the likeness and fine-scale details of the original subject, but a perceptual refinement stage is necessary to transfer person-specific facial characteristics from the input photo. Our experiments also show that perceptual loss enables more robust matching using deep features than only pixel loss, and is able to better preserve semantically meaningful facial features.
Compared to state-of-the-art non-linear 3DMMs, our generated face models can produce lighting and expression normalized face models, which is a requirement for seamless integration of avatars in virtual environments. Furthermore, our experiments also indicate that our results are not only perceptually superior, but also quantitatively more accurate and robust than existing methods.

\paragraph{Limitations and Future Work.} As shown in Fig.~\ref{fig:failure_case}, the effectiveness of our method in generating faces with normalized expressions and lighting is limited by imperfect training data and challenging input photos. In particular, some expressions and specularities can still be found in the generated results.
Furthermore, the fundamental problem of disentangling identity from expressions, or lighting conditions from skin tones is ill-posed. Nevertheless, we believe that such disentanglement can be improved using superior training data. In the future, we would like to explore how to increase the resolution and fidelity of the digitized assets and potentially combine our method with high-fidelity facial asset inference techniques such as~\cite{Lattas_2020_CVPR,chen2019photo,Yamaguchi_2018}.

{\small
\bibliographystyle{ieee_fullname}
\bibliography{egbib}

\begin{thebibliography}{10}\itemsep=-1pt

\bibitem{3D_scan_store}
3d scan store.
\newblock \url{https://www.3dscanstore.com/}.

\bibitem{tripelgangers}
Triplegangers.
\newblock \url{http://www.triplegangers.com}.

\bibitem{Abdal_2019_ICCV}
Rameen Abdal, Yipeng Qin, and Peter Wonka.
\newblock Image2stylegan: How to embed images into the stylegan latent space?
\newblock In {\em Proceedings of the IEEE/CVF International Conference on
  Computer Vision (ICCV)}, October 2019.

\bibitem{Abdal_2020_CVPR}
Rameen Abdal, Yipeng Qin, and Peter Wonka.
\newblock Image2stylegan++: How to edit the embedded images?
\newblock In {\em IEEE/CVF Conference on Computer Vision and Pattern
  Recognition (CVPR)}, June 2020.

\bibitem{Abrevaya_2019_CVPR}
Victoria~Fern{\'{a}}ndez Abrevaya, Adnane Boukhayma, Stefanie Wuhrer, and
  Edmond Boyer.
\newblock A decoupled 3d facial shape model by adversarial training.
\newblock In {\em CVPR}, 2019.

\bibitem{Bagautdinov_2018_CVPR}
Timur Bagautdinov, Chenglei Wu, Jason Saragih, Pascal Fua, and Yaser Sheikh.
\newblock Modeling facial geometry using compositional vaes.
\newblock In {\em CVPR}, 2018.

\bibitem{Bas_2017_ICCV_Workshops}
Anil Bas, Patrik Huber, William A.~P. Smith, Muhammad Awais, and Josef Kittler.
\newblock 3d morphable models as spatial transformer networks.
\newblock In {\em Proceedings of the IEEE International Conference on Computer
  Vision (ICCV) Workshops}, Oct 2017.

\bibitem{Beeler_2010_SIGGRAPH}
Thabo Beeler, Bernd Bickel, Paul Beardsley, Bob Sumner, and Markus Gross.
\newblock High-quality single-shot capture of facial geometry.
\newblock {\em {ACM} Trans. Graph.}, 29(4), 2010.

\bibitem{blanz2003reanimating}
Volker Blanz, Curzio Basso, Tomaso Poggio, and Thomas Vetter.
\newblock Reanimating faces in images and video.
\newblock In {\em Computer graphics forum}, volume~22. Wiley Online Library,
  2003.

\bibitem{Blanz_1999_SIGGRAPH}
Volker Blanz and Thomas Vetter.
\newblock A morphable model for the synthesis of 3d faces.
\newblock In {\em Proceedings of the 26th Annual Conference on Computer
  Graphics and Interactive Techniques}, SIGGRAPH ’99, page 187–194, USA,
  1999. ACM Press/Addison-Wesley Publishing Co.

\bibitem{Booth_2017_CVPR}
James Booth, Epameinondas Antonakos, Stylianos Ploumpis, George Trigeorgis,
  Yannis Panagakis, and Stefanos Zafeiriou.
\newblock 3d face morphable models "in-the-wild".
\newblock In {\em CVPR}, 2017.

\bibitem{Booth_2016_CVPR}
James Booth, Anastasios Roussos, Stefanos Zafeiriou, Allan Ponniah, and David
  Dunaway.
\newblock A 3d morphable model learnt from 10,000 faces.
\newblock In {\em CVPR}, 2016.

\bibitem{Bouaziz:2013:OMR}
Sofien Bouaziz, Yangang Wang, and Mark Pauly.
\newblock Online modeling for realtime facial animation.
\newblock {\em ACM Trans. Graph.}, 32(4), 2013.

\bibitem{brock2018large}
Andrew Brock, Jeff Donahue, and Karen Simonyan.
\newblock Large scale {GAN} training for high fidelity natural image synthesis.
\newblock In {\em International Conference on Learning Representations}, 2019.

\bibitem{Cao:2014:DDE}
Chen Cao, Qiming Hou, and Kun Zhou.
\newblock Displaced dynamic expression regression for real-time facial tracking
  and animation.
\newblock {\em ACM Trans. Graph.}, 33(4), 2014.

\bibitem{cao2014facewarehouse}
Chen Cao, Yanlin Weng, Shun Zhou, Yiying Tong, and Kun Zhou.
\newblock Facewarehouse: A 3d facial expression database for visual computing.
\newblock {\em IEEE TVCG}, 20(3), 2014.

\bibitem{cao2016real}
Chen Cao, Hongzhi Wu, Yanlin Weng, Tianjia Shao, and Kun Zhou.
\newblock Real-time facial animation with image-based dynamic avatars.
\newblock {\em ACM Trans. Graph.}, 35(4), 2016.

\bibitem{chen2019photo}
Anpei Chen, Zhang Chen, Guli Zhang, Kenny Mitchell, and Jingyi Yu.
\newblock Photo-realistic facial details synthesis from single image.
\newblock In {\em CVPR}, 2019.

\bibitem{Cheng_2019_MeshGAN}
Shiyang Cheng, Michael~M. Bronstein, Yuxiang Zhou, Irene Kotsia, Maja Pantic,
  and Stefanos Zafeiriou.
\newblock Meshgan: Non-linear 3d morphable models of faces.
\newblock {\em CoRR}, 2019.

\bibitem{cole2017synthesizing}
Forrester Cole, David Belanger, Dilip Krishnan, Aaron Sarna, Inbar Mosseri, and
  William~T Freeman.
\newblock Synthesizing normalized faces from facial identity features.
\newblock In {\em Proceedings of the IEEE conference on computer vision and
  pattern recognition}, pages 3703--3712, 2017.

\bibitem{deng2019accurate}
Yu Deng, Jiaolong Yang, Sicheng Xu, Dong Chen, Yunde Jia, and Xin Tong.
\newblock Accurate 3d face reconstruction with weakly-supervised learning: From
  single image to image set.
\newblock In {\em IEEE Computer Vision and Pattern Recognition Workshops},
  2019.

\bibitem{Dou_2017_CVPR}
Pengfei Dou, Shishir~K. Shah, and Ioannis~A. Kakadiaris.
\newblock End-to-end 3d face reconstruction with deep neural networks.
\newblock In {\em Proceedings of the IEEE Conference on Computer Vision and
  Pattern Recognition (CVPR)}, July 2017.

\bibitem{du2014compound}
Shichuan Du, Yong Tao, and Aleix~M Martinez.
\newblock Compound facial expressions of emotion.
\newblock {\em Proceedings of the National Academy of Sciences},
  111(15):E1454--E1462, 2014.

\bibitem{feng2018prn}
Yao Feng, Fan Wu, Xiaohu Shao, Yanfeng Wang, and Xi Zhou.
\newblock Joint 3d face reconstruction and dense alignment with position map
  regression network.
\newblock In {\em ECCV}, 2018.

\bibitem{Fyffe_2016_EG}
G. Fyffe, P. Graham, B. Tunwattanapong, A. Ghosh, and P. Debevec.
\newblock Near-instant capture of high-resolution facial geometry and
  reflectance.
\newblock In {\em Proceedings of the 37th Annual Conference of the European
  Association for Computer Graphics}, EG ’16, page 353–363, Goslar, DEU,
  2016. Eurographics Association.

\bibitem{GVWT13}
Pablo Garrido, Levi Valgaerts, Chenglei Wu, and Christian Theobalt.
\newblock Reconstructing detailed dynamic face geometry from monocular video.
\newblock In {\em {ACM} Trans. Graph. (Proceedings of SIGGRAPH Asia 2013)},
  volume~32, November 2013.

\bibitem{Garrido_2016_SIGGRAPH}
Pablo Garrido, Michael Zollh{\"o}fer, Dan Casas, Levi Valgaerts, Kiran
  Varanasi, Patrick Perez, and Christian Theobalt.
\newblock Reconstruction of personalized 3d face rigs from monocular video.
\newblock {\em {ACM} Trans. Graph. (Presented at SIGGRAPH 2016)}, 35(3), 2016.

\bibitem{Gecer_2019_CVPR}
Baris Gecer, Stylianos Ploumpis, Irene Kotsia, and Stefanos Zafeiriou.
\newblock Ganfit: Generative adversarial network fitting for high fidelity 3d
  face reconstruction.
\newblock In {\em Proceedings of the IEEE/CVF Conference on Computer Vision and
  Pattern Recognition (CVPR)}, June 2019.

\bibitem{Genova_2018_CVPR}
Kyle Genova, Forrester Cole, Aaron Maschinot, Aaron Sarna, Daniel Vlasic, and
  William~T. Freeman.
\newblock Unsupervised training for 3d morphable model regression.
\newblock In {\em Proceedings of the IEEE Conference on Computer Vision and
  Pattern Recognition (CVPR)}, June 2018.

\bibitem{ghosh2011multiview}
Abhijeet Ghosh, Graham Fyffe, Borom Tunwattanapong, Jay Busch, Xueming Yu, and
  Paul Debevec.
\newblock Multiview face capture using polarized spherical gradient
  illumination.
\newblock volume~30, page 129. ACM, 2011.

\bibitem{NIPS2014_5423}
Ian Goodfellow, Jean Pouget-Abadie, Mehdi Mirza, Bing Xu, David Warde-Farley,
  Sherjil Ozair, Aaron Courville, and Yoshua Bengio.
\newblock Generative adversarial nets.
\newblock In {\em Advances in Neural Information Processing Systems 27}, pages
  2672--2680. 2014.

\bibitem{GROSS2010807}
Ralph Gross, Iain Matthews, Jeffrey Cohn, Takeo Kanade, and Simon Baker.
\newblock Multi-pie.
\newblock {\em Image and Vision Computing}, 28(5):807 -- 813, 2010.
\newblock Best of Automatic Face and Gesture Recognition 2008.

\bibitem{guan2020collaborative}
Shanyan Guan, Ying Tai, Bingbing Ni, Feida Zhu, Feiyue Huang, and Xiaokang
  Yang.
\newblock Collaborative learning for faster stylegan embedding, 2020.

\bibitem{He_2016_CVPR}
Kaiming He, Xiangyu Zhang, Shaoqing Ren, and Jian Sun.
\newblock Deep residual learning for image recognition.
\newblock In {\em Proceedings of the IEEE Conference on Computer Vision and
  Pattern Recognition (CVPR)}, June 2016.

\bibitem{hsieh2015unconstrained}
Pei-Lun Hsieh, Chongyang Ma, Jihun Yu, and Hao Li.
\newblock Unconstrained realtime facial performance capture.
\newblock In {\em IEEE CVPR}, 2015.

\bibitem{Hu_2017_SIGGRAPHASIA}
Liwen Hu, Shunsuke Saito, Lingyu Wei, Koki Nagano, Jaewoo Seo, Jens Fursund,
  Iman Sadeghi, Carrie Sun, Yen-Chun Chen, and Hao Li.
\newblock Avatar digitization from a single image for real-time rendering.
\newblock {\em ACM Trans. Graph.}, Nov. 2017.

\bibitem{Ichim:2015:DAC}
Alexandru~Eugen Ichim, Sofien Bouaziz, and Mark Pauly.
\newblock Dynamic 3d avatar creation from hand-held video input.
\newblock {\em ACM Trans. Graph.}, 34(4), 2015.

\bibitem{karras2018progressive}
Tero Karras, Timo Aila, Samuli Laine, and Jaakko Lehtinen.
\newblock Progressive growing of {GAN}s for improved quality, stability, and
  variation.
\newblock In {\em International Conference on Learning Representations}, 2018.

\bibitem{Karras_2019_CVPR}
Tero Karras, Samuli Laine, and Timo Aila.
\newblock A style-based generator architecture for generative adversarial
  networks.
\newblock In {\em Proceedings of the IEEE/CVF Conference on Computer Vision and
  Pattern Recognition (CVPR)}, June 2019.

\bibitem{Karras_2020_CVPR}
Tero Karras, Samuli Laine, Miika Aittala, Janne Hellsten, Jaakko Lehtinen, and
  Timo Aila.
\newblock Analyzing and improving the image quality of {StyleGAN}.
\newblock In {\em Proceedings of the IEEE/CVF Conference on Computer Vision and
  Pattern Recognition (CVPR)}, 2020.

\bibitem{Ira_2013_ICCV}
Ira Kemelmacher-Shlizerman.
\newblock Internet-based morphable model.
\newblock {\em ICCV}, 2013.

\bibitem{doi:10.1080/02699930903485076}
Oliver Langner, Ron Dotsch, Gijsbert Bijlstra, Daniel H.~J. Wigboldus,
  Skyler~T. Hawk, and Ad van Knippenberg.
\newblock Presentation and validation of the radboud faces database.
\newblock {\em Cognition and Emotion}, 24(8):1377--1388, 2010.

\bibitem{Lattas_2020_CVPR}
Alexandros Lattas, Stylianos Moschoglou, Baris Gecer, Stylianos Ploumpis,
  Vasileios Triantafyllou, Abhijeet Ghosh, and Stefanos Zafeiriou.
\newblock Avatarme: Realistically renderable 3d facial reconstruction
  "in-the-wild".
\newblock In {\em CVPR}, June 2020.

\bibitem{CelebAMask-HQ}
Cheng-Han Lee, Ziwei Liu, Lingyun Wu, and Ping Luo.
\newblock Maskgan: Towards diverse and interactive facial image manipulation.
\newblock In {\em IEEE Conference on Computer Vision and Pattern Recognition
  (CVPR)}, 2020.

\bibitem{Lee_2020_CVPR}
Gun-Hee Lee and Seong-Whan Lee.
\newblock Uncertainty-aware mesh decoder for high fidelity 3d face
  reconstruction.
\newblock In {\em Proceedings of the IEEE/CVF Conference on Computer Vision and
  Pattern Recognition (CVPR)}, June 2020.

\bibitem{li2013realtime}
Hao Li, Jihun Yu, Yuting Ye, and Chris Bregler.
\newblock Realtime facial animation with on-the-fly correctives.
\newblock {\em ACM Trans. Graph. (Proceedings SIGGRAPH 2013)}, 32(4), 2013.

\bibitem{li2020learning}
Ruilong Li, Karl Bladin, Yajie Zhao, Chinmay Chinara, Owen Ingraham, Pengda
  Xiang, Xinglei Ren, Pratusha Prasad, Bipin Kishore, Jun Xing, et~al.
\newblock Learning formation of physically-based face attributes.
\newblock In {\em CVPR}, 2020.

\bibitem{Li_2017_SIGGRAPHASIA}
Tianye Li, Timo Bolkart, Michael~J. Black, Hao Li, and Javier Romero.
\newblock Learning a model of facial shape and expression from 4d scans.
\newblock {\em ACM Trans. Graph.}, 36(6), Nov. 2017.

\bibitem{Lin_2020_CVPR}
Jiangke Lin, Yi Yuan, Tianjia Shao, and Kun Zhou.
\newblock Towards high-fidelity 3d face reconstruction from in-the-wild images
  using graph convolutional networks.
\newblock In {\em IEEE/CVF Conference on Computer Vision and Pattern
  Recognition (CVPR)}, June 2020.

\bibitem{hifi3dface2020tencentailab}
Xiangkai Lin, Yajing Chen, Linchao Bao, Haoxian Zhang, Sheng Wang, Xuefei Zhe,
  Xinwei Jiang, Jue Wang, Dong Yu, and Zhengyou Zhang.
\newblock High-fidelity 3d digital human creation from rgb-d selfies.
\newblock {\em arXiv preprint arXiv:2010.05562}, 2020.

\bibitem{Litany_2018_CVPR}
Or Litany, Alex Bronstein, Michael Bronstein, and Ameesh Makadia.
\newblock Deformable shape completion with graph convolutional autoencoders.
\newblock In {\em CVPR}, 2018.

\bibitem{Ma_2015_data}
Debbie Ma, Joshua Correll, and Bernd Wittenbrink.
\newblock The chicago face database: A free stimulus set of faces and norming
  data.
\newblock {\em Behavior research methods}, 47:1122--1135, 01 2015.

\bibitem{Nagano_2019_Siggraph}
Koki Nagano, Huiwen Luo, Zejian Wang, Jaewoo Seo, Jun Xing, Liwen Hu, Lingyu
  Wei, and Hao Li.
\newblock Deep face normalization.
\newblock {\em ACM Trans. Graph.}, 2019.

\bibitem{paszke2019pytorch}
Adam Paszke, Sam Gross, Francisco Massa, Adam Lerer, James Bradbury, Gregory
  Chanan, Trevor Killeen, Zeming Lin, Natalia Gimelshein, Luca Antiga, et~al.
\newblock Pytorch: An imperative style, high-performance deep learning library.
\newblock In {\em Advances in neural information processing systems}, pages
  8026--8037, 2019.

\bibitem{Ranjan_2018_ECCV}
Anurag Ranjan, Timo Bolkart, Soubhik Sanyal, and Michael~J. Black.
\newblock Generating 3d faces using convolutional mesh autoencoders.
\newblock In {\em ECCV}, 2018.

\bibitem{ravi2020pytorch3d}
Nikhila Ravi, Jeremy Reizenstein, David Novotny, Taylor Gordon, Wan-Yen Lo,
  Justin Johnson, and Georgia Gkioxari.
\newblock Accelerating 3d deep learning with pytorch3d.
\newblock {\em arXiv:2007.08501}, 2020.

\bibitem{Sami_2005_CVPR}
Romdhani Sami and Thomas Vetter.
\newblock Estimating 3d shape and texture using pixel intensity, edges,
  specular highlights, texture constraints and a prior.
\newblock In {\em Proceedings of the IEEE Conference on Computer Vision and
  Pattern Recognition (CVPR)}, 2005.

\bibitem{Schroff_2015_CVPR}
Florian Schroff, Dmitry Kalenichenko, and James Philbin.
\newblock Facenet: A unified embedding for face recognition and clustering.
\newblock In {\em Proceedings of the IEEE Conference on Computer Vision and
  Pattern Recognition (CVPR)}, June 2015.

\bibitem{stylegan2-pytorch}
Kim Seonghyeon.
\newblock stylegan2-pytorch.
\newblock \url{https://github.com/rosinality/stylegan2-pytorch}.

\bibitem{Shen_2020_CVPR}
Yujun Shen, Jinjin Gu, Xiaoou Tang, and Bolei Zhou.
\newblock Interpreting the latent space of gans for semantic face editing.
\newblock In {\em IEEE/CVF Conference on Computer Vision and Pattern
  Recognition (CVPR)}, June 2020.

\bibitem{Shi:2014:AAH}
Fuhao Shi, Hsiang-Tao Wu, Xin Tong, and Jinxiang Chai.
\newblock Automatic acquisition of high-fidelity facial performances using
  monocular videos.
\newblock {\em ACM Trans. Graph.}, 33(6), 2014.

\bibitem{shi2019probabilistic}
Yichun Shi and Anil~K Jain.
\newblock Probabilistic face embeddings.
\newblock In {\em Proceedings of the IEEE International Conference on Computer
  Vision}, pages 6902--6911, 2019.

\bibitem{Tewari_2017_ICCV}
Ayush Tewari, Michael Zollhofer, Hyeongwoo Kim, Pablo Garrido, Florian Bernard,
  Patrick Perez, and Christian Theobalt.
\newblock Mofa: Model-based deep convolutional face autoencoder for
  unsupervised monocular reconstruction.
\newblock In {\em ICCV}, 2017.

\bibitem{Thies_2016_CVPR}
Justus Thies, Michael Zollhöfer, Marc Stamminger, Christian Theobalt, and
  Matthias Nießner.
\newblock Face2face: Real-time face capture and reenactment of rgb videos.
\newblock In {\em CVPR}, 2016.

\bibitem{Tran_2019_CVPR}
Luan Tran, Feng Liu, and Xiaoming Liu.
\newblock Towards high-fidelity nonlinear 3d face morphable model.
\newblock In {\em CVPR}, 2019.

\bibitem{Tran_2018_CVPR}
Luan Tran and Xiaoming Liu.
\newblock Nonlinear 3d face morphable model.
\newblock In {\em CVPR}, 2018.

\bibitem{Tran_2017_CVPR}
Anh Tuan~Tran, Tal Hassner, Iacopo Masi, and Gerard Medioni.
\newblock Regressing robust and discriminative 3d morphable models with a very
  deep neural network.
\newblock In {\em Proceedings of the IEEE Conference on Computer Vision and
  Pattern Recognition (CVPR)}, July 2017.

\bibitem{Vlasic:2005:FTM}
Daniel Vlasic, Matthew Brand, Hanspeter Pfister, and Jovan Popovi\'{c}.
\newblock Face transfer with multilinear models.
\newblock {\em ACM Trans. Graph.}, 24(3), 2005.

\bibitem{weise2011realtime}
Thibaut Weise, Sofien Bouaziz, Hao Li, and Mark Pauly.
\newblock Realtime performance-based facial animation.
\newblock {\em ACM Trans. Graph. (Proceedings SIGGRAPH 2011)}, 30(4), July
  2011.

\bibitem{weise09face}
Thibaut Weise, Hao Li, Luc~Van Gool, and Mark Pauly.
\newblock Face/off: Live facial puppetry.
\newblock In {\em Proceedings of the 2009 ACM SIGGRAPH/Eurographics Symposium
  on Computer animation (Proc. SCA'09)}, ETH Zurich, August 2009. Eurographics
  Association.

\bibitem{wu2019mvf}
Fanzi Wu, Linchao Bao, Yajing Chen, Yonggen Ling, Yibing Song, Songnan Li,
  King~Ngi Ngan, and Wei Liu.
\newblock Mvf-net: Multi-view 3d face morphable model regression.
\newblock In {\em CVPR}, 2019.

\bibitem{Yamaguchi_2018}
Shugo Yamaguchi, Shunsuke Saito, Koki Nagano, Yajie Zhao, Weikai Chen, Kyle
  Olszewski, Shigeo Morishima, and Hao Li.
\newblock High-fidelity facial reflectance and geometry inference from an
  unconstrained image.
\newblock {\em ACM Trans. Graph.}, 37(4), 2018.

\bibitem{yang2020facescape}
Haotian Yang, Hao Zhu, Yanru Wang, Mingkai Huang, Qiu Shen, Ruigang Yang, and
  Xun Cao.
\newblock Facescape: a large-scale high quality 3d face dataset and detailed
  riggable 3d face prediction.
\newblock In {\em Proceedings of the IEEE Conference on Computer Vision and
  Pattern Recognition (CVPR)}, 2020.

\bibitem{zhang2018perceptual}
Richard Zhang, Phillip Isola, Alexei~A Efros, Eli Shechtman, and Oliver Wang.
\newblock The unreasonable effectiveness of deep features as a perceptual
  metric.
\newblock In {\em CVPR}, 2018.

\bibitem{zhao2017pspnet}
Hengshuang Zhao, Jianping Shi, Xiaojuan Qi, Xiaogang Wang, and Jiaya Jia.
\newblock Pyramid scene parsing network.
\newblock In {\em CVPR}, 2017.

\bibitem{Zhou_2019_CVPR}
Yuxiang Zhou, Jiankang Deng, Irene Kotsia, and Stefanos Zafeiriou.
\newblock Dense 3d face decoding over 2500fps: Joint texture \& shape
  convolutional mesh decoders.
\newblock In {\em CVPR}, 2019.

\bibitem{zhu2019lia}
Jiapeng Zhu, Deli Zhao, Bo Zhang, and Bolei Zhou.
\newblock Disentangled inference for gans with latently invertible autoencoder.
\newblock {\em arXiv preprint arXiv:1906.08090}, 2019.

\end{thebibliography}
}

\newpage
\section*{Appendix I. Additional Comparisons}

\begin{figure}[hbt!]
 \includegraphics[width=3.25in]{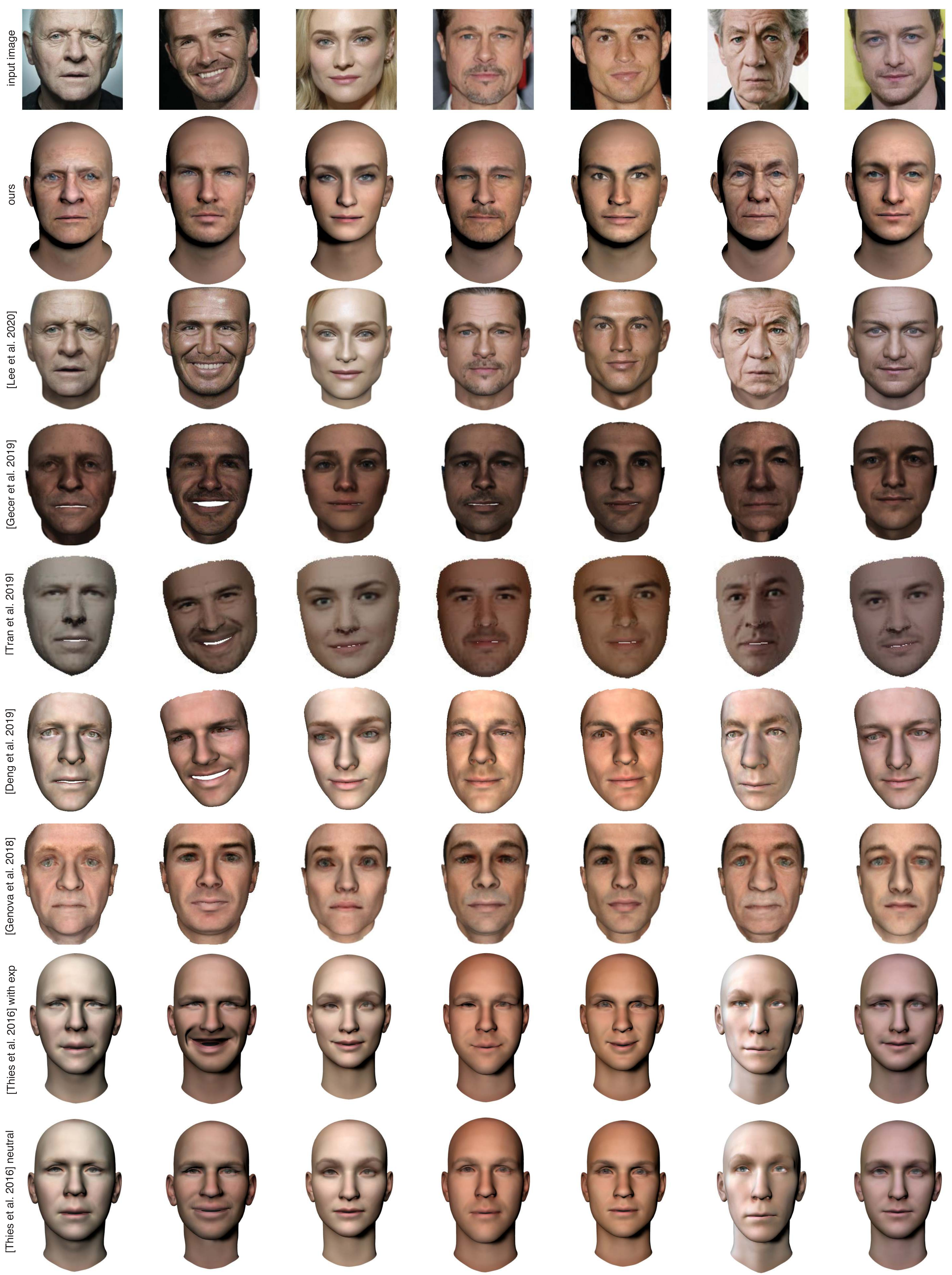}
 \caption{Additional comparisons. The first row shows the input images and the second row our results. The remaining rows are the reconstructed 3D faces obtained by \cite{Lee_2020_CVPR,Gecer_2019_CVPR,Tran_2019_CVPR,deng2019accurate,Genova_2018_CVPR,Thies_2016_CVPR}, respectively.}
 \label{fig:comparison_supp}
\end{figure}

In Fig.~\ref{fig:comparison_supp}, we compare our method with several recent state-of-the-art single view face reconstruction approaches. Thies et al. ~\cite{Thies_2016_CVPR} extend the seminal work of Blanz and Vetter~\cite{Blanz_1999_SIGGRAPH} with facial expression blendshapes and iteratively optimize for shape, texture, and lighting condition by minimizing energy terms based on facial landmark and pixel color constraints. We visualize the avatars with and without the facial expressions of the corresponding input photo. Neutralizing facial expressions is straightforward by setting all the blendshape coefficients to $0$.
We notice that the linear morphable face model is unable to recover features such as facial hair, as well as high-frequency geometry and appearance details. As a result, the face renderings often lack the likeness of the original subject and often fall within the so called ``uncanny valley''.
Genova et al. ~\cite{Genova_2018_CVPR} predict identity coefficients of linear 3DMM using a deep neural network and Deng et al.~\cite{deng2019accurate} predict the lights and face poses simultaneously using additionally linear 3DMM coefficients. Their models are still restricted to the linear subspace which has limited capabilities for representing facial details. Gecer et al.~\cite{Gecer_2019_CVPR} introduce an unsupervised training approach to regress linear 3DMM coefficients for geometry and adopt a Generative Adversarial Network model for generating nonlinear texture. Tran et al.~\cite{Tran_2019_CVPR} present an approach to learn additional proxies as means to avoid strong regularization, which efficiently captures high level details for geometry and texture with a simple decoder architecture. They do not separate identity and expressions in the training. Lee et al.~\cite{Lee_2020_CVPR} demonstrate the latest work for generating 3D face models from a single input photograph using non-linear 3DMMs and an uncertainty-aware mesh decoder. The resulting 3D faces are very faithful to the input image, but the lighting and expressions are baked into the texture and mesh. As a result, neither Lee et al.~\cite{Lee_2020_CVPR} nor the above non-linear 3DMM techniques produce normalized results as shown in our paper. Notice that the results in Fig.~\ref{fig:comparison_supp} from row 3 to row 7 were taken directly from the paper of ~\cite{Lee_2020_CVPR}, and the renderings may have slight inconsistencies.

\section*{Appendix II. Additional Evaluations}

\begin{figure}[hbt!]
 \includegraphics[width=3.25in]{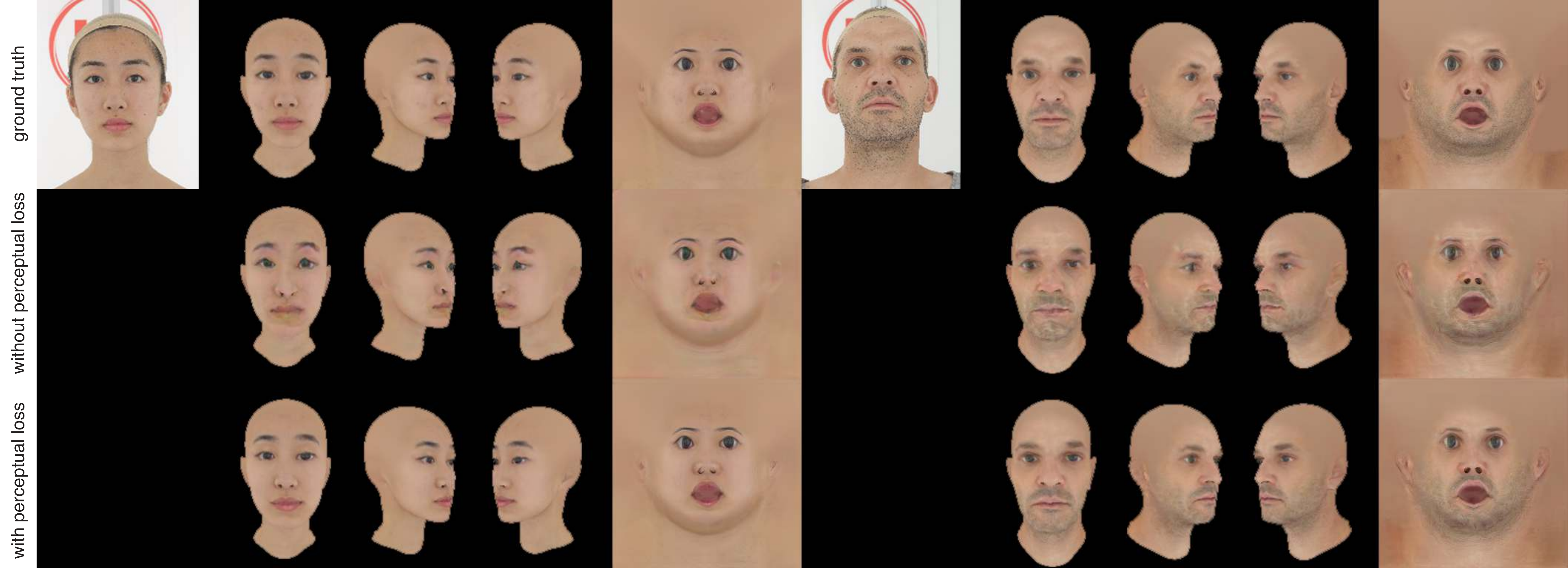}
 \caption{Algorithmic choice justification on the loss function for GAN-inversion. From top to bottom: Ground truth geometry and texture; Reconstruction results optimized by pixel loss and adversarial loss; Reconstruction results with perceptual loss in addition.}
 \label{fig:projection_loss}
\end{figure}

In Sec. 3.1, we adopt a two step training method by first training $G$ and then freezing $G$ in order to compute the code inversion and to train $R$. %The key assumptions of such a method is that $G$ can embed the nonlinear 3DMM into the smooth latent space without any identity feature knowledge, and the code inversion method can reliably recover the corresponding latent codes. %We try to validate these assumptions via Fig.~\ref{fig:styles_interpolation} and Fig.~\ref{fig:projection_loss}. Fig.~\ref{fig:styles_interpolation} presents examples of 3D faces synthesized by mixing latent vectors at various scales, similar to StyleGAN2. We can see that each subset of styles controls meaningful high-level attributes of the identity, which is the characteristic of StyleGAN2, and $G$ indeed embeds the nonlinear 3DMM of the position and texture map successfully, since the latent codes can be freely interpolated without causing undesired artifacts.
Fig.~\ref{fig:projection_loss} shows that the latent codes can be effectively found out with our choice of loss function in Eq. 2. Specifically, while pixel loss and adversarial loss cannot preserve the overall similarity, adding the perceptual loss improves the high-level appearance in the rendering views.

\begin{figure}[hbt!]
 \includegraphics[width=3.25in]{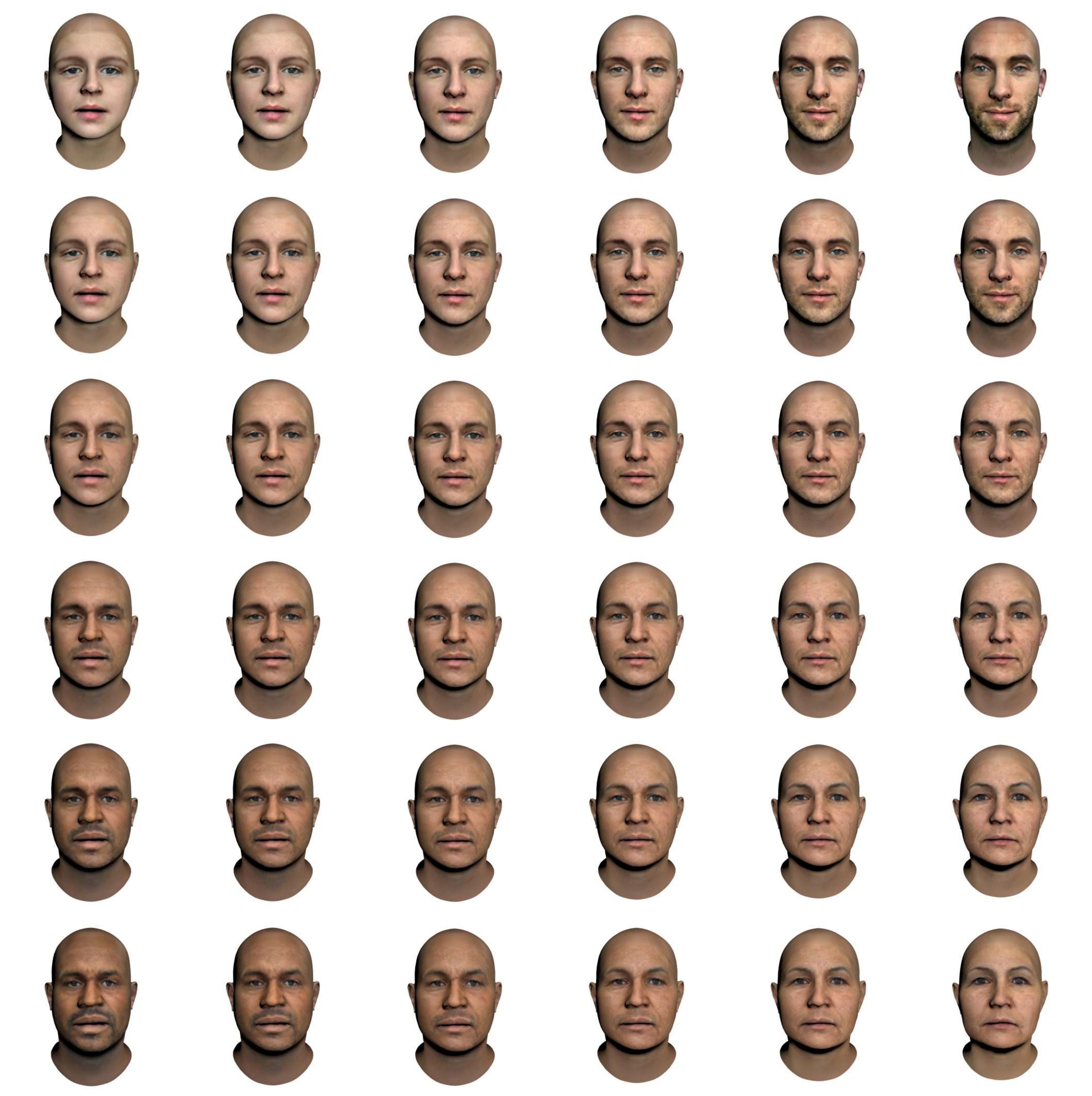}
 \caption{Illustration of latent vector interpolation. The four input 3D avatars are shown
at the corners, while all the in-between interpolations are based on
bi-linear interpolated weights.}
\label{fig:interpolation}
\end{figure}

\paragraph{Face Interpolation.}
In Fig.~\ref{fig:interpolation}, we show interpolation results of multiple 3D avatars.
The four input avatars are shown at the corners. All the interpolation results are obtained via bi-linearly interpolation of the embedding $\textbf{w}$ computed from the four images.
As shown in the results, realistic, plausible, and artifact free avatar assets can be generated using our method, which can be useful for a wide range of avatar manipulation and synthesis tasks.

\begin{figure}[hbt!]
 \includegraphics[width=3.25in]{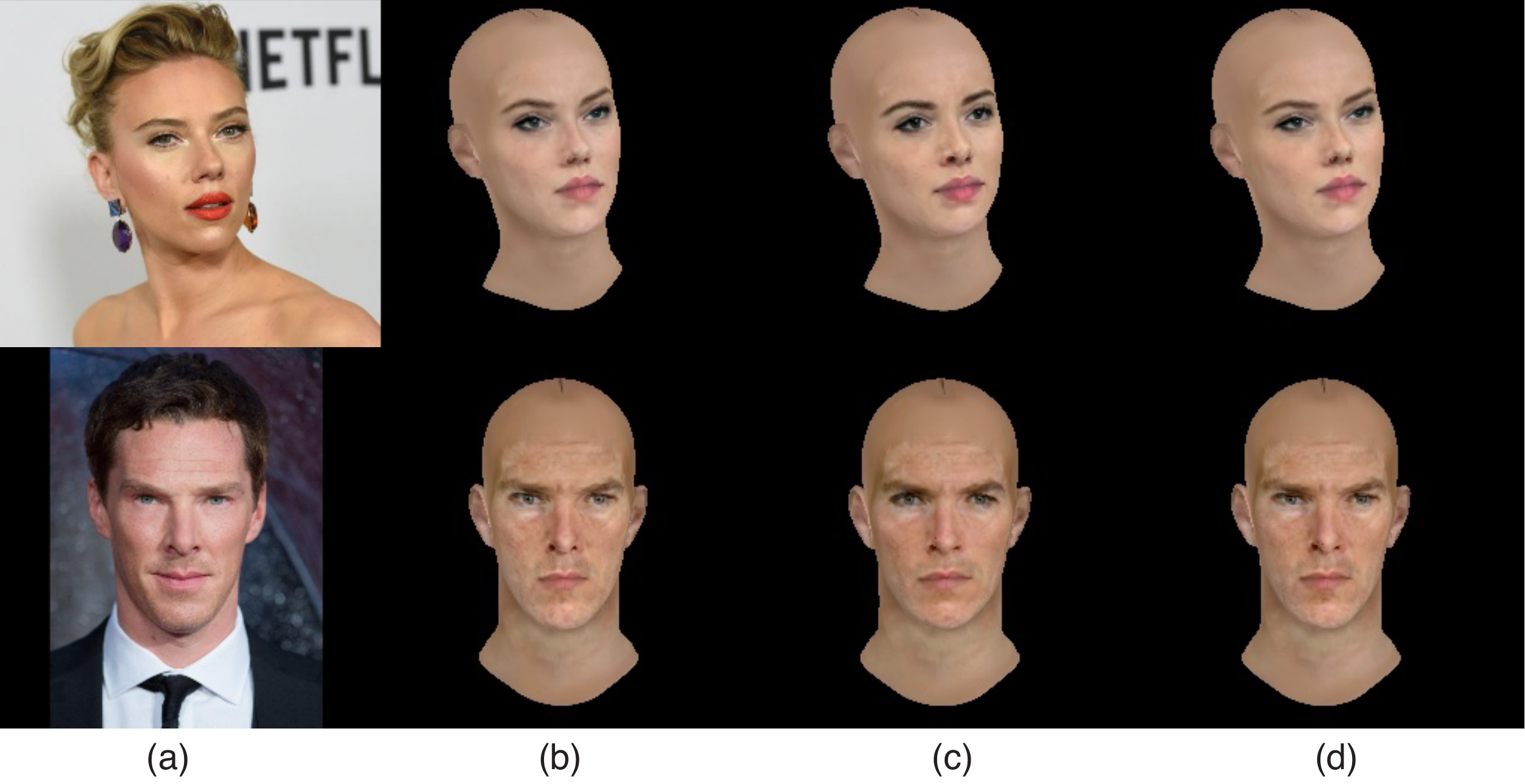}
 \caption{Visual comparison illustrating the effects of losses in the perceptual refinement step, where the full model leads to better results. From left to right: (a) input image; (b) refinement result with identity loss and $\textbf{w}$ regularization; (c) refinement result with perceptual loss and $\textbf{w}$ regularization; (d) refinement result with all three losses.}
 \label{fig:optimization_loss}
\end{figure}

\paragraph{Optimization Loss.}
Fig.~\ref{fig:optimization_loss} shows the benefit of each loss term in $L_{refine}$ for the perceptual refinement. Combining identity loss, perceptual loss, and $\textbf{w}$ regularization allows us to generate clean assets, where the resulting subject preserves the likeness of the subject in the original input photo, but at the same time, ensures consistent and detailed assets with normalized lighting and neutral expressions.

\begin{figure}[hbt!]
 \includegraphics[width=3.25in]{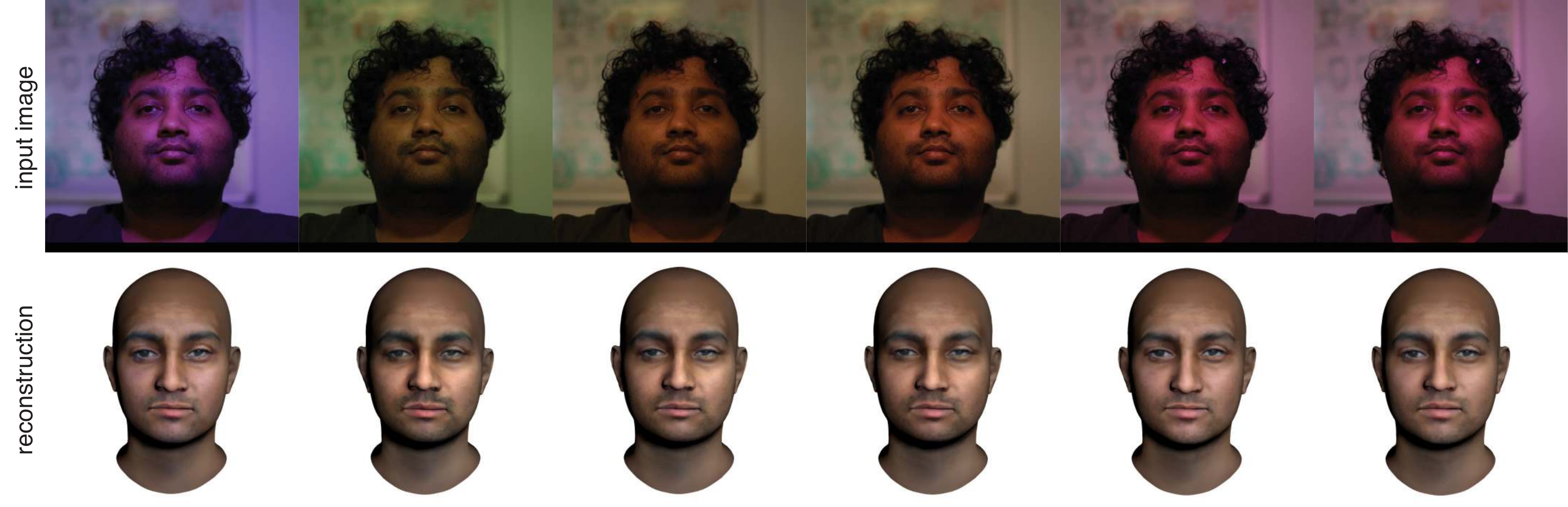}
 \caption{Consistent reconstructions of albedo texture under varying extreme illuminations.}
 \label{fig:illuminations_consistency}
\end{figure}

\paragraph{Illumination Consistency.}
Fig.~\ref{fig:illuminations_consistency} demonstrates consistent face reconstructions of albedo textures from varying illuminations conditions. In this experiment we move around a light with different extreme colors around the subjects and demonstrate how a consistent 3D avatar with a nearly identical dark skin tone is correctly reconstructed for each input photo.

\begin{figure}[hbt!]
 \includegraphics[width=3.25in]{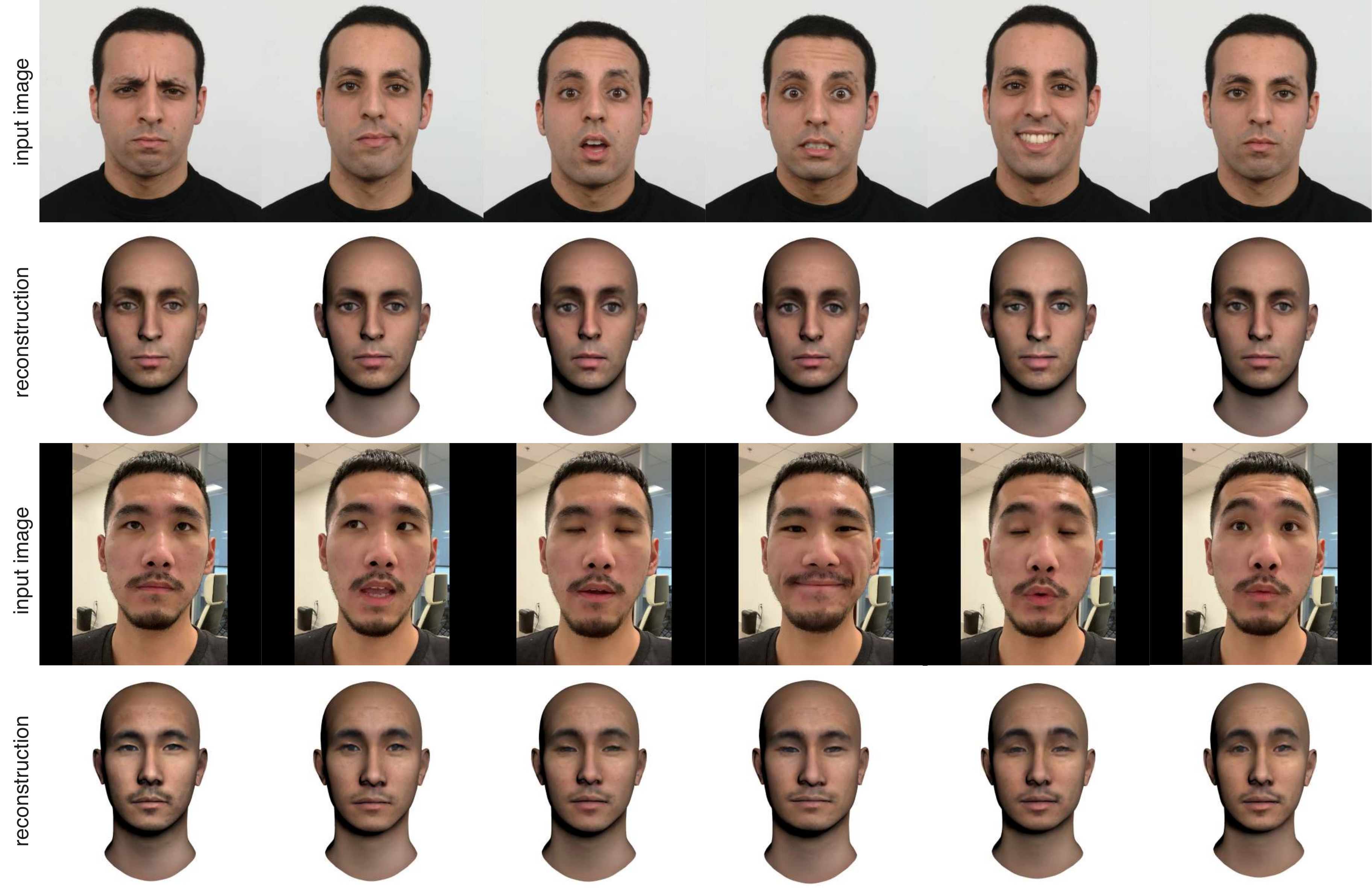}
 \caption{Consistent reconstructions of 3D avatars from images with different expressions.}
 \label{fig:expression_consistency}
\end{figure}

\paragraph{Expression Consistency.}
We demonstrate how consistent faces are reconstructed from input images with different expressions in Fig.~\ref{fig:expression_consistency}. In particular, our method digitizes consistent 3D avatars with neutral expressions despite a wide range of diverse and extreme facial expressions of the same person as shown in the first row and the third row. While some amount of the input expressions are reflected in the normalized results, the overall neutralization is significantly superior than existing techniques, especially for extreme input facial expressions. 

\begin{figure}[hbt]
 \centering
 \includegraphics[width=3.25in]{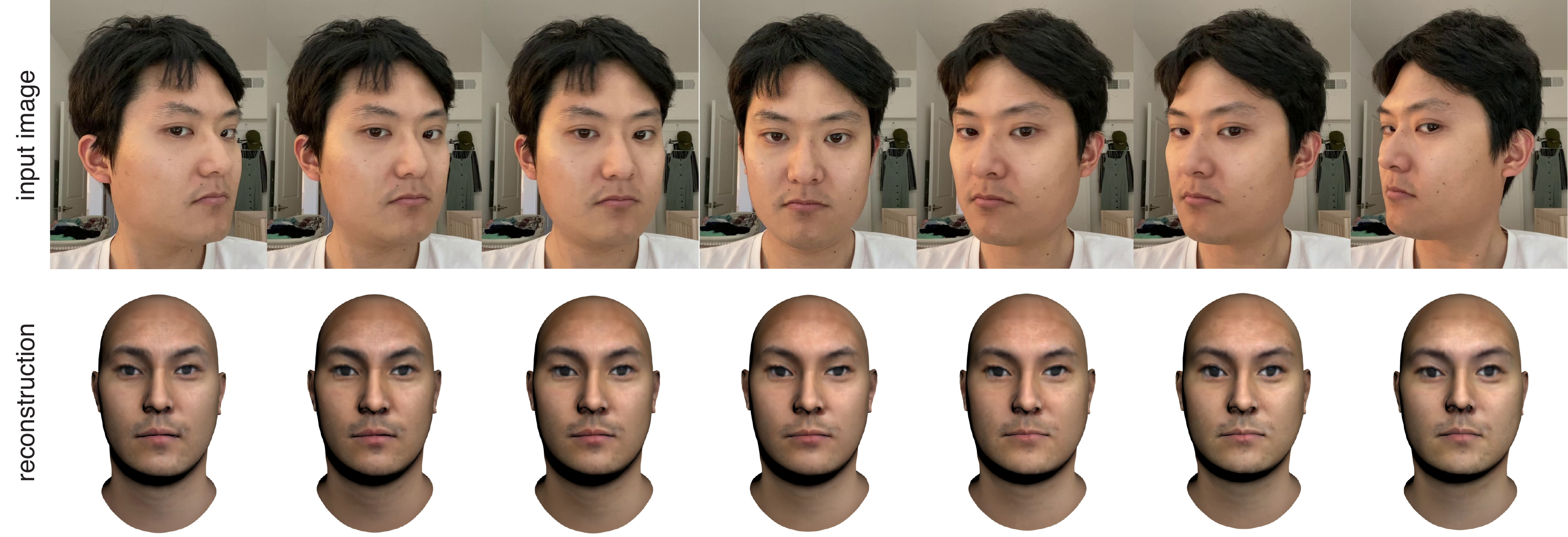}
 \caption{Consistent reconstructions under different poses.}
 \label{fig:pose_consistency}
\end{figure}

\paragraph{Pose Consistency.}
Fig.~\ref{fig:pose_consistency} shows consistent reconstructions from varying head poses. For side views, our method can still generate highly consistent textures and geometries despite non-visible face regions in the input image.

\section*{Appendix III. Additional Results}

\begin{figure*}[hbt!]
\centering
 \includegraphics[width=6.75in]{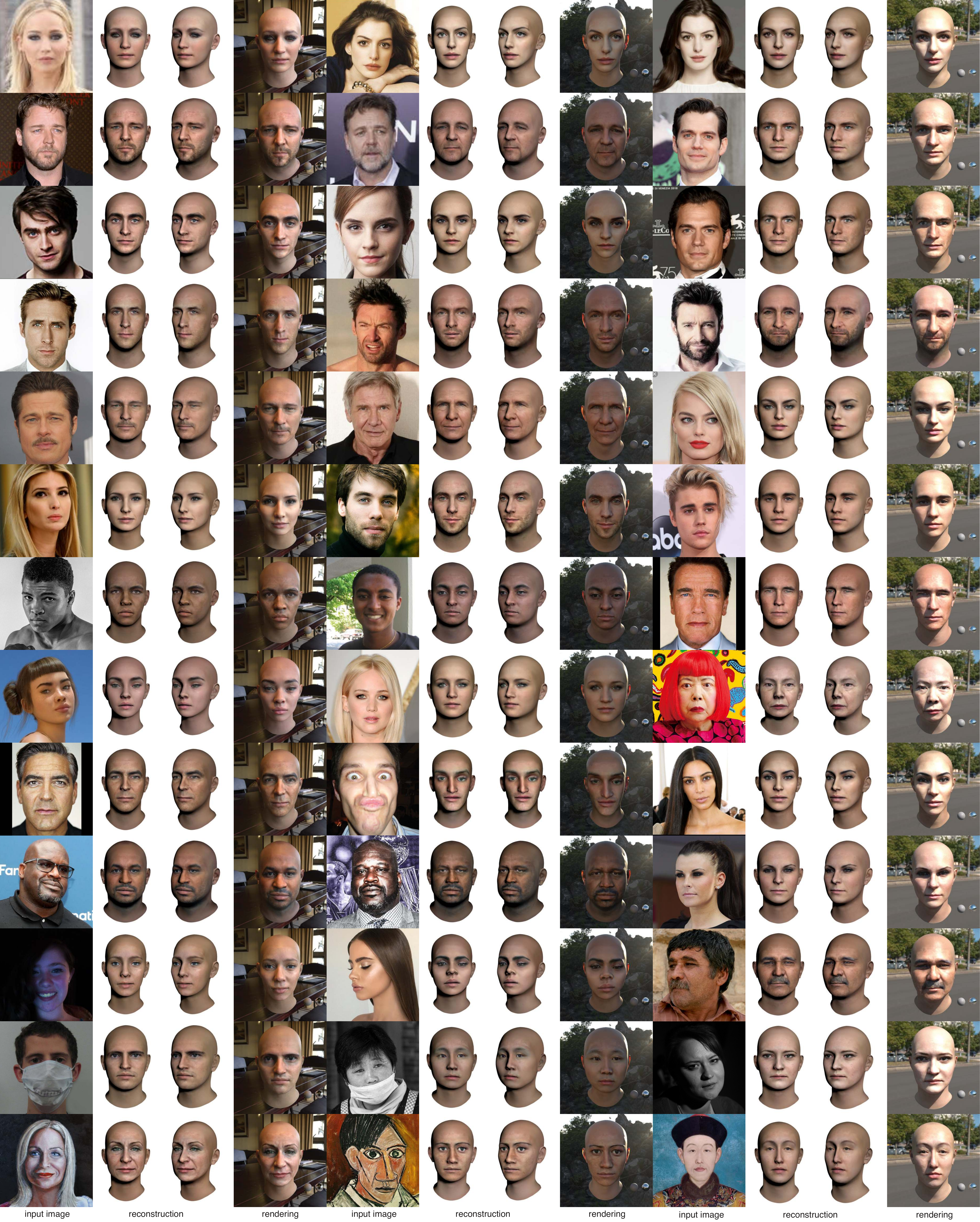}
 \caption{Batch 1 additional results of normalized 3D avatars from a single input image. None of these subjects have been used in training for our networks.}
 \label{fig:more_results1}
\end{figure*}

\begin{figure*}[hbt!]
\centering
 \includegraphics[width=6.75in]{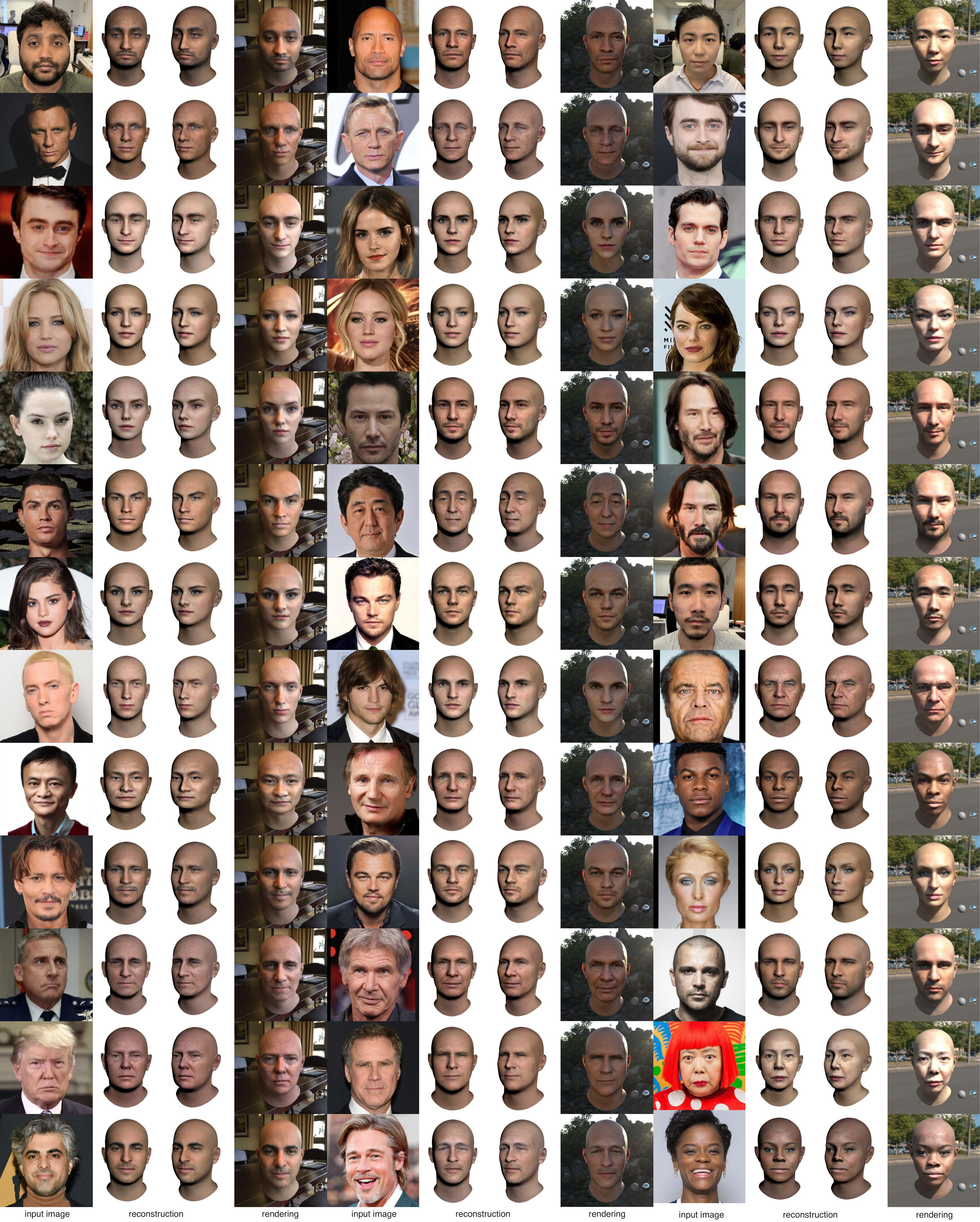}
 \caption{Batch 2 additional results of normalized 3D avatars from a single input image. None of these subjects have been used in training for our networks.}
 \label{fig:more_results2}
\end{figure*}

\begin{figure*}[hbt!]
\centering
 \includegraphics[width=6.75in]{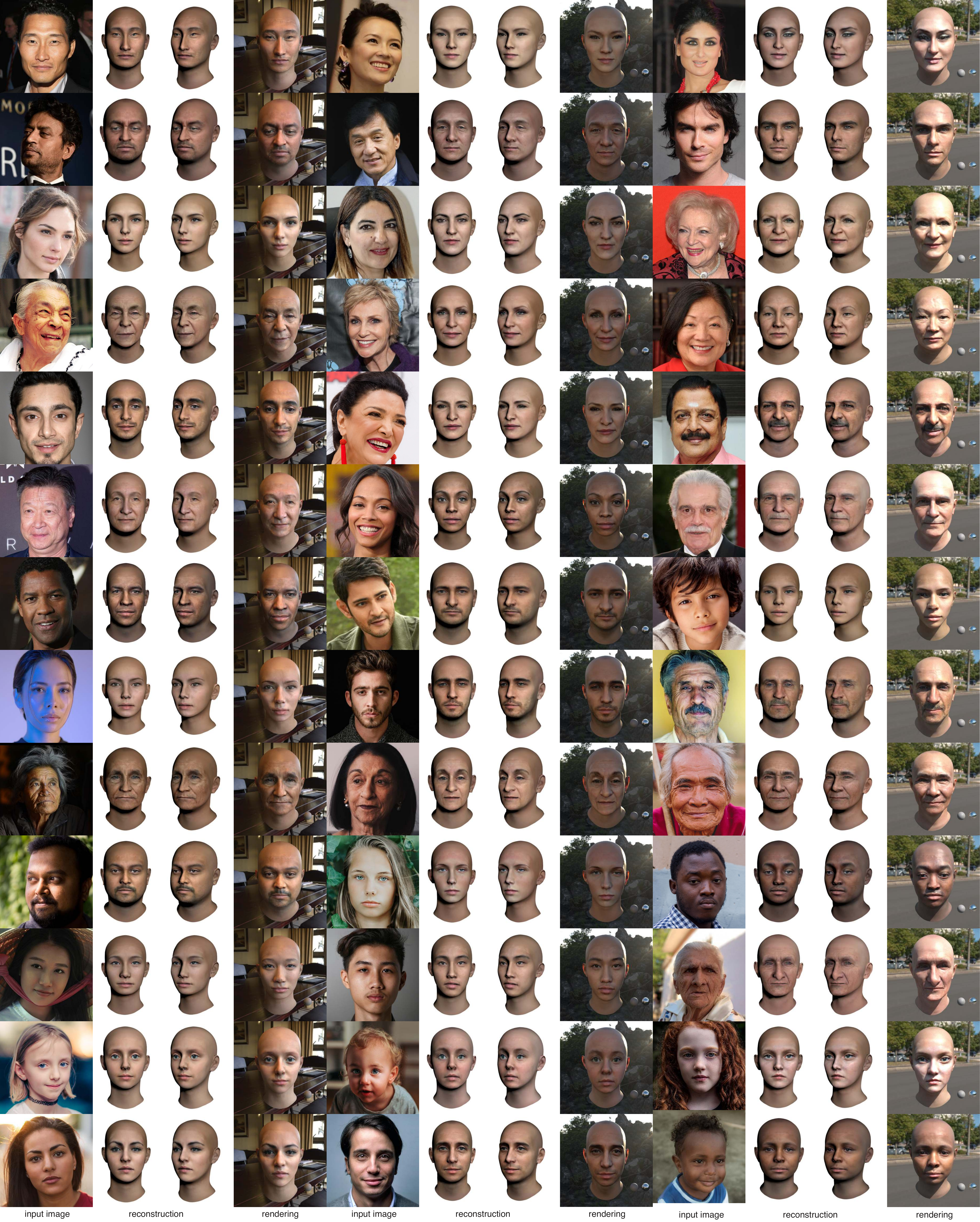}
 \caption{Batch 3 additional results of normalized 3D avatars from a single input image. None of these subjects have been used in training for our networks.}
 \label{fig:more_results3}
\end{figure*}

\begin{figure*}[hbt!]
\centering
 \includegraphics[width=6.75in]{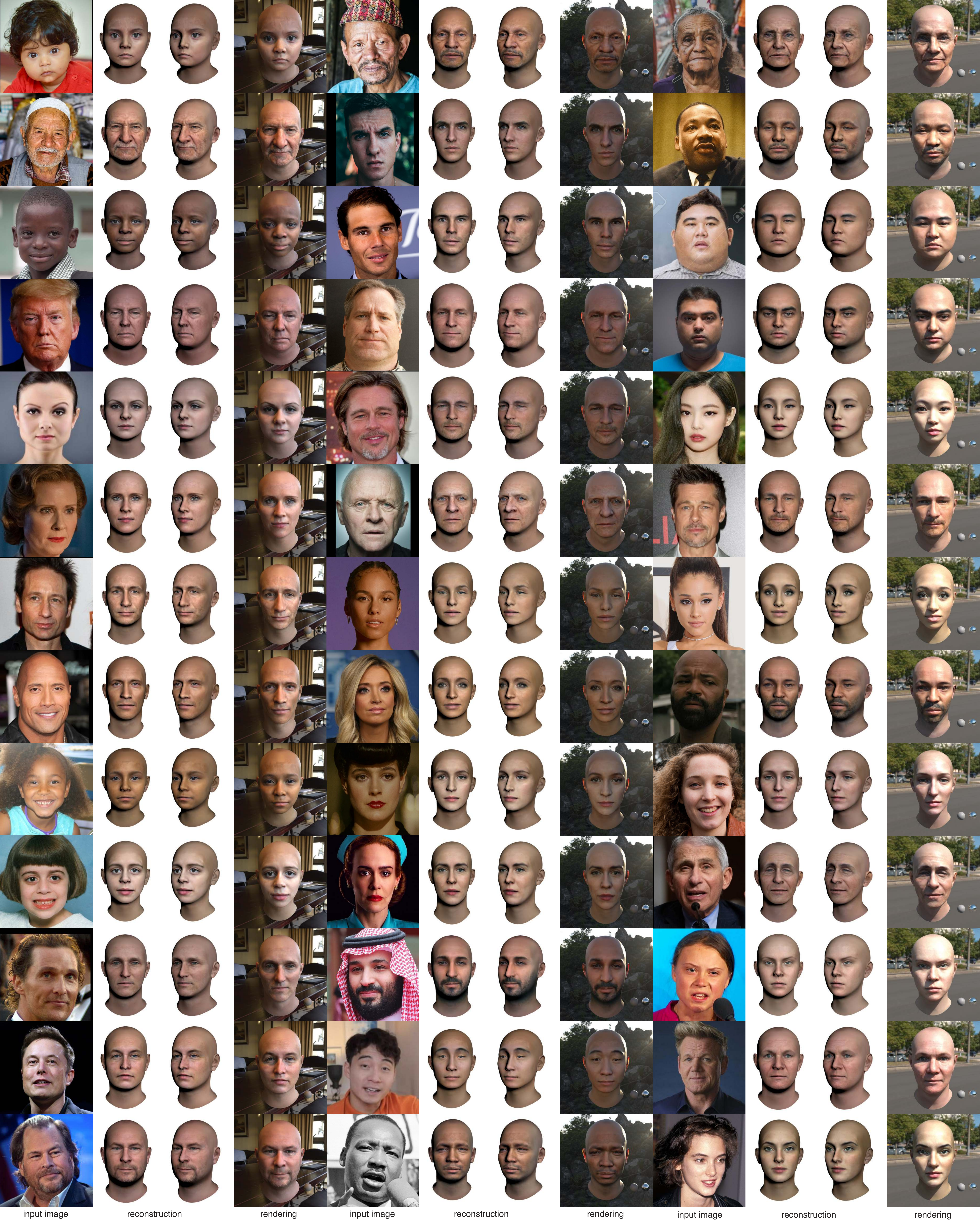}
 \caption{Batch 4 additional results of normalized 3D avatars from a single input image. None of these subjects have been used in training for our networks.}
 \label{fig:more_results4}
\end{figure*}

To demonstrate the robustness of the our technique, we provide $156$ additional examples with a wider range of extremely challenging input photographs in Fig.~\ref{fig:more_results1}, Fig.~\ref{fig:more_results2}, Fig.~\ref{fig:more_results3}, and Fig.~\ref{fig:more_results4}. These figures illustrate input pictures, successful normalized 3D face reconstructions, as well as renderings using HDRI-based lighting environments. Our results include diverse ethnicity, both genders, and varying age groups, ranging from children to old people. We also showcase a wide range of complex lighting conditions, stylized photographs, black and white portraits, drawings and paintings, facial occlusions, as well as a wide range of extreme head poses and facial expressions. Notice that we also show several results of the same person, but reconstructed from entirely different input images.

\end{document}